%% file: jodogne07a.tex
\documentclass[twoside,11pt,letterpaper]{article}
\input epsf 

\usepackage{jair}
\usepackage{theapa}

\jairheading{28}{2007}{349--391}{06/06}{03/07}
\ShortHeadings{Closed-Loop Learning of Visual Control Policies}
{Jodogne \& Piater}
\firstpageno{349}

\input{definitions}

\begin{document}

\title{Closed-Loop Learning of Visual Control Policies}

\author{\name S{\'e}bastien Jodogne \email Jodogne@Montefiore.ULg.ac.be \\
       \name Justus H. Piater \email Justus.Piater@ULg.ac.be \\
       \addr Montefiore Institute (B28) \\
       University of Liège, B-4000 Li{\`e}ge, Belgium}


\maketitle

\begin{abstract}
\input{abstract}
\end{abstract}

\input{introduction}
\input{reinforcement-learning}
\input{rlvc}

\input{compacting}

\input{spatial}
\input{conclusions}

\vskip 0.2in
\bibliography{bibtex/jodogne,bibtex/aliasing,bibtex/reinforcement-learning,bibtex/vision,bibtex/machine-learning,bibtex/neuropsychology,bibtex/divers,bibtex/piater,bibtex/minimization}
\bibliographystyle{theapa}

\end{document}

%% file: definitions.tex
\usepackage{graphicx,url}
\usepackage{color}

\usepackage[latin1]{inputenc}
\usepackage{nofontwarning}
\usepackage{amsfonts}
\usepackage{amssymb}
\usepackage{amsmath}
\usepackage{url}
\usepackage[ruled]{algorithm}
\usepackage{algorithmic}

\let\ifFTR\iffalse

\def\Rset{\mathbb{R}}
\def\argmax{\mathop{\rm argmax}}
\def\argmin{\mathop{\rm argmin}}

\def\mean{\mathop{\rm mean}}
\def\stddev{\mathop{\rm stddev}}

%% file: abstract.tex
In this paper we present a general, flexible framework for learning
mappings from images to actions by interacting with the environment.
The basic idea is to introduce a feature-based image classifier in
front of a reinforcement learning algorithm. The classifier partitions
the visual space according to the presence or absence of few highly
informative local descriptors that are incrementally selected in a
sequence of attempts to remove perceptual aliasing. We also address
the problem of fighting overfitting in such a greedy
algorithm. Finally, we show how high-level visual features can be
generated when the power of local descriptors is insufficient for
completely disambiguating the aliased states. This is done by building
a hierarchy of composite features that consist of recursive spatial
combinations of visual features. We demonstrate the efficacy of our
algorithms by solving three visual navigation tasks and a visual
version of the classical ``Car on the Hill'' control problem.

%% file: introduction.tex
\section{Introduction}

Designing robotic controllers quickly becomes a challenging
problem. Indeed, such controllers face a huge number of possible
inputs that can be noisy, must select actions among a continuous set,
and should be able to automatically adapt themselves to evolving or
stochastic environmental conditions. Although a real-world robotic
task can often be solved by directly connecting the perceptual space
to the action space through a given computational mechanism, such
mappings are usually hard to derive by hand, especially when the
perceptual space contains images. Evidently, automatic methods for
generating such mappings are highly desirable, because many robots are
nowadays equipped with CCD sensors.

In this paper, we are interested in reactive systems that learn to
couple visual perceptions and actions inside a dynamic world so as to
act reasonably. This coupling is known as a {\it visual (control)
policy\/}.  This wide category of problems will be called {\it
vision-for-action tasks\/} (or simply {\it visual tasks\/}).  Despite
about fifty years years of active research in artificial intelligence,
robotic agents are still largely unable to solve many real-world
visuomotor tasks that are easily performed by humans and even by
animals. Such vision-for-action tasks notably include grasping,
vision-guided navigation and manipulation of objects so as to achieve
a goal. This article introduces a general framework that is suitable
for building image-to-action mappings using a fully automatic and
flexible learning protocol.

\subsection{Vision-for-Action and Reinforcement Learning}

Strong neuropsychological evidence suggests that human beings learn to
extract useful information from visual data in an {\it interactive\/}
fashion, without any external supervisor~\cite{Gibson:DP83}.  By
evaluating the consequence of our actions on the environment, we learn
to pay attention to visual cues that are behaviorally important for
solving the task. This way, as we interact with the outside world, we
gain more and more expertise on our tasks~\cite{Tarr:TCS2003}.
Obviously, this process is {\it task driven\/}, since different tasks
do not necessarily need to make the same
distinctions~\cite{Schyns:JEPLMC1997}. 

A breakthrough in modern artificial intelligence would be to design an
artificial system that would acquire object or scene recognition
skills based only on its experience with the surrounding
environment. To state it in more general terms, an important research
direction would be to design a robotic agent that could {\it
autonomously acquire visual skills from its interactions\/} with an
uncommitted environment in order to achieve some set of
goals. Learning new visual skills in a dynamic, task-driven fashion so
as to complete an {\it a priori\/} unknown visual task is known as the
{\it purposive vision\/} paradigm~\cite{Aloimonos:ICPR1990}.

One plausible framework to learn image-to-action mappings according to
purposive vision is {\it Reinforcement Learning\/}
(RL)~\cite{Bertsekas:book-1996,Kaelbling:JAIR1996,Sutton:book-1998}.
Reinforcement learning is a biologically-inspired computational
framework that can generate nearly optimal control policies in an
automatic way, by interacting with the environment. RL is founded on
the analysis of a so-called {\it reinforcement signal\/}.  Whenever
the agent takes a decision, it receives as feedback a real number that
evaluates the relevance of this decision. From a biological
perspective, when this signal becomes positive, the agent experiences
pleasure, and we can talk about a {\it reward\/}. Conversely, a
negative reinforcement implies a sensation of pain, which corresponds
to a {\it punishment\/}. The reinforcement signal can be arbitrarily
delayed from the actions which are responsible for it.  Now, RL
algorithms are able to map every possible perception to an action that
maximizes the reinforcement signal {\it over time\/}. In this
framework, the agent is never told what the optimal action is when
facing a given percept, nor whether one of its decisions was
optimal. Rather, the agent has to discover by itself what the most
promising actions are by constituting a representative database of
interactions, and by understanding the influence of its decisions on
future reinforcements. Schematically, RL lies between supervised
learning (where an external teacher gives the correct action to the
agent) and unsupervised learning (in which no clue about the goodness
of the action is given).

RL has had successful applications, for example turning a computer
into an excellent Backgammon player~\cite{Tesauro:CACM1995}, solving
the Acrobot control problem~\cite{Yoshimoto:ICSMC-1999}, making a
quadruped robot learn progressively to walk without any human
intervention~\cite{Huber:ISORA1998,Kimura:CDC2001,Kohl:ICRA2004},
riding a bicycle~\cite{Randlov:ICML1998,Lagoudakis:JMLR2003} or
controlling a helicopter~\cite{Bagnell:ICRA2001,Ng:ISER2004}. The
major advantages of the RL protocol are that it is fully automatic,
and that it imposes very weak constraints on the environment.

Unfortunately, standard RL algorithms are highly sensitive to the
number of distinct percepts as well as to the noise that results from
the sensing process. This general problem is often referred to as the
{\it Bellman curse of
dimensionality\/}~\cite{Bellman:book-1957}. Thus, the high
dimensionality and the noise that is inherent to images forbid the use
of basic RL algorithms for direct closed-loop learning of
image-to-action mappings according to purposive vision.

\subsection{Achieving Purposive Vision through Reinforcement Learning}

There exists a variety of work in RL on specific robotic problems
involving a perceptual space that contains images. For instance,
Schaal~\citeyear{Schaal:ANIPS1997} uses visual feedback to solve a
pole-balancing task. RL has been used to control a vision-guided
underwater robotic vehicle~\cite{Wettergreen:ICFSR1999}.  More
recently, Kwok and Fox~\citeyear{Kwok:IROS2004} have demonstrated the
applicability of RL to learning sensing strategies using {\sc Aibo}
robots. Reinforcement learning can also be used to learn strategies
for view selection~\cite{Paletta:RAS2000} and sequential attention
models~\cite{Paletta:ICML2005}. Let us also mention the use of
reinforcement learning in other vision-guided tasks such as ball
kicking~\cite{Asada:WVB1994}, ball
acquisition~\cite{Takahashi:ICMFIIS1999}, visual
servoing~\cite{Gaskett:ACRA2000}, robot
docking~\cite{Weber:KBS2004,Martinez:ICRA2005} and obstacle
avoidance~\cite{Michels:ICML2005}. Interestingly, RL is also used as a
way of tuning the high-level parameters of image-processing
applications. For example, Peng and Bhanu~\citeyear{Peng:PAMI1998}
introduce RL algorithms for image segmentation, whereas
Yin~\citeyear{Yin:SP2002} proposes algorithms for multilevel image
thresholding, and uses entropy as a reinforcement signal.

All of these applications preprocess the images to extract some
high-level information about the observed scene that is directly
relevant to the task to be solved and that feeds the RL algorithm.
This requires prior assumptions about the images perceived by the
sensors of the agent, and about the physical structure of the task
itself. The preprocessing step is task specific and is coded by
hand. This contrasts with our objectives, which consist in introducing
algorithms able to {\it learn\/} how to {\it directly\/} connect the
visual space to the action space, without using manually written code
and without relying on prior knowledge about the task to be
solved. Our aim is to develop general algorithms that are applicable
to any visual task that can be formulated in the RL framework.

A noticeable exception is the work by Iida et
al.~\citeyear{Iida:ICONIP2002} who apply RL to seek and reach targets,
and to push boxes~\cite{Shibata:SICE2003} with real robots. In this
work, raw visual signals directly feed a neural network that is
trained by an actor-critic architecture. In these examples, the visual
signal is downscaled and averaged into a monochrome (i.e.~two-color)
image of $64\times 24=1536$ pixels. The output of four infrared
sensors are also added to this perceptual input. While this approach
is effective for the specific tasks, this process can only be used in
a highly controlled environment. Real-world images are much richer and
could not undergo such a strong reduction in size.

\subsection{Local-Appearance Paradigm}
\label{local-appearance}

In this paper, we propose algorithms that rely on the {\it extraction
of visual features\/} as a way to achieve more compact state spaces
that can be used as an input to traditional RL algorithms. Indeed,
buried in the noise and in the confusion of visual cues, images
contain hints of regularity.  Such regularities are captured by the
important notion of {\it visual features\/}. Loosely speaking, a
visual feature is a representation of some aspect of local appearance,
e.g.~a corner formed by two intensity edges, a spatially localized
texture signature, or a color. Therefore, to analyze images, it is
often sufficient for a computer program to extract only useful
information from the visual signal, by focusing its attention on
robust and highly informative patterns in the percepts. The program
should thereafter seek the characteristic appearance of the observed
scenes or objects.

This is actually the basic postulate behind {\it local-appearance
methods\/} that have had much success in computer vision applications
such as image matching, image retrieval and object
recognition~\cite{Schmid:PAMI1997,Lowe:IJCV2004}.  They rely on the
detection of discontinuities in the visual signal thanks to {\it
interest point detectors\/}~\cite{Schmid:IJCV2000}. Similarities in
images are thereafter identified using a {\it local description\/} of
the neighborhood around the interest
points~\cite{Mikolajczyk:CVPR2003}: If two images share a sufficient
number of matching local descriptors, they are considered to belong to
the same visual class.

\begin{figure*}
\begin{center}
\includegraphics[height=3cm]{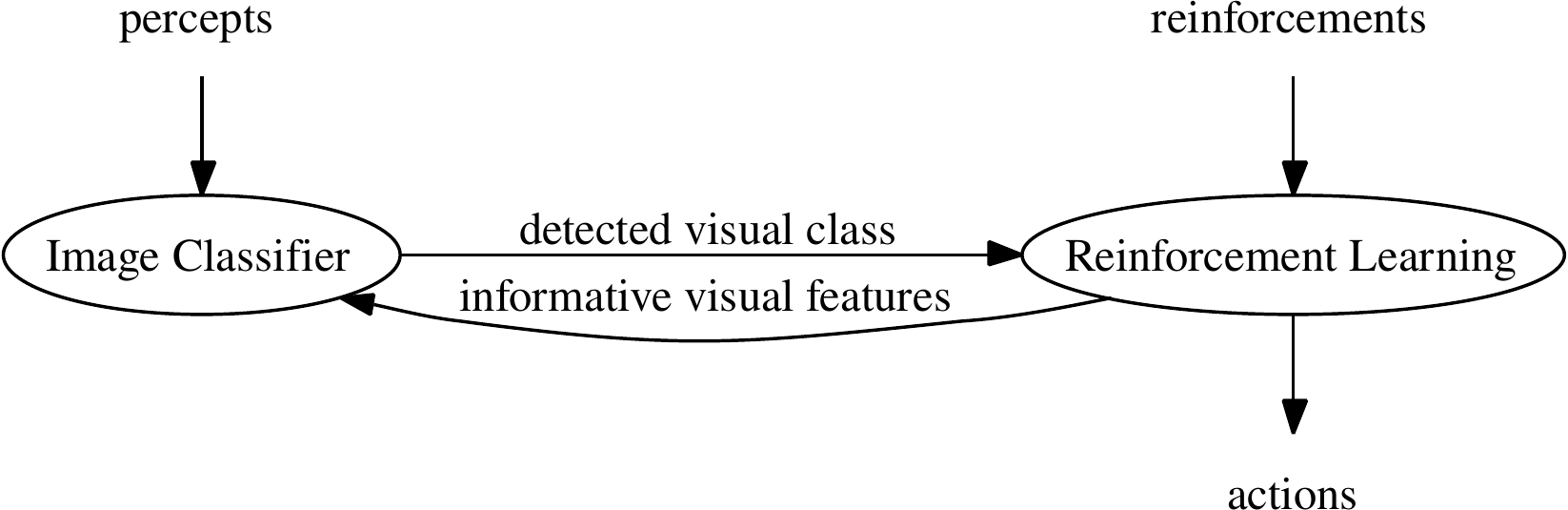}
\end{center}
\caption{The structure of Reinforcement Learning of Visual
Classes.\label{fig-system}}
\end{figure*}

Local-appearance techniques are at the same time powerful and
flexible, as they are robust to partial occlusions, and do not require
segmentation or 3D models of the scenes. It seems therefore promising
to introduce, in front of the RL algorithm, a feature-based {\it image
classifier\/} that partitions the visual space into a finite set of
distinct {\it visual classes\/} according to the local-appearance
paradigm, by focusing the attention of the agent on highly distinctive
local descriptors located at interest points of the visual
stimuli. The symbol corresponding to the detected visual class could
then be given as the input of a classical, embedded RL algorithm, as
shown in Figure~\ref{fig-system}.

This preprocessing step is intended to reduce the size of the input
domain, thus enhancing the rate of convergence, the generalization
capabilities as well as the robustness of RL to noise in visual
domains. Importantly, the same family of visual features can be
applied to a wide variety of visual tasks, thus the preprocessing step
is essentially general and task-independent.  The central difficulty
is the dynamic selection of the discriminative visual features. This
selection process should group images that share similar,
task-specific properties together in the same visual class.

\subsection{Contributions}

The key technical contribution of this paper consists in the
introduction of reinforcement learning algorithms that can be used
when the perceptual space contains images. The developed algorithms do
not rely on a task-specific pre-treatment. As a consequence, they can
be used in any vision-for-action task that can be formalized as a
Markov Decision Problem. We now review the three major contributions
that are discussed in this paper.

\subsubsection{Adaptive Discretization of a Visual Space}

Our first contribution is to propose a new algorithm called {\it
Reinforcement Learning of Visual Classes\/} (RLVC) that combines the
aforementioned ideas. RLVC is an iterative algorithm that is suitable
for learning direct image-to-action mappings by taking advantage of
the local-appearance paradigm. It consists of two simultaneous,
interleaved learning processes: Reinforcement learning of a mapping
from visual classes to actions, and incremental building of a
feature-based image classifier.

Initially, the image classifier contains one single visual class, so
that all images are mapped to this class. Of course, this introduces a
kind of {\it perceptual aliasing\/} (or {\it hidden
state\/})~\cite{Whitehead:ML-1991}: The optimal decisions cannot
always be made, since percepts requiring different reactions are
associated with the same class. The agent then isolates the aliased
classes. Since there is no external supervisor, the agent can only
rely on a statistical analysis of the earned reinforcements. For each
detected aliased class, the agent dynamically selects a new visual
feature that is {\it distinctive\/}, i.e.~that best disambiguates the
aliased percepts. The extracted local descriptor is used to refine the
classifier. This way, at each stage of the algorithm, the number of
visual classes in the classifier grows.  New visual features are
learned until perceptual aliasing vanishes. The resulting image
classifier is finally used to control the system.

Our approach is primarily motivated by strong positive results of
McCallum's {\it U-Tree\/} algorithm~\cite{McCallum:phd-1996}. In
essence, RLVC is an adaptation of U-Tree to visual spaces, though the
internals of the algorithms are different. The originality of RLVC
lies in its exploitation of successful local-appearance features. RLVC
selects a subset of such highly relevant features in a fully
closed-loop, purposive learning process. We show that this algorithm
is of practical interest, as it can be successfully applied to several
simulated visual navigation tasks.

\subsubsection{Compacting Visual Policies}
\label{intro-compact}

Because of its greedy nature, RLVC is prone to overfitting. Splitting
{\it one\/} visual class can potentially improve the control policy
for {\it all\/} the visual classes. Therefore, the splitting strategy
can get stuck in local minima: Once a split is made that subsequently
proves useless, it cannot be undone in the original description of
RLVC. Our second contribution is to provide RLVC with the possibility
of aggregating visual classes that share similar properties. Doing so
has at least three potential benefits:
\begin{enumerate}
\item Useless features are discarded, which enhances generalization
capabilities;
\item RLVC can reset the search for good features; and
\item the number of samples that the embedded RL algorithm has at its
disposal for each visual class is increased, which results in better
visual control policies.
\end{enumerate}
Experiments indeed show an improvement in the generalization
abilities, as well as a reduction of the number of visual classes and
selected features.

\subsubsection{Spatial Combinations of Visual Features}
\label{intro-spatial}

Finally, the efficacy of RLVC clearly depends on the discriminative
power of the visual features. If their power is insufficient, the
algorithm will not be able to completely remove the aliasing, which
will produce sub-optimal control policies. Practical experiments on
simulated visual navigation tasks exhibit this deficiency, as soon as
the number of detected visual features is reduced or as features are
made more similar by using a less sensitive metric.  Now, most objects
encountered in the world are composed of a number of distinct
constituent parts (e.g.~a face contains a nose and two eyes, a phone
possesses a keypad). These parts are themselves recursively composed
of other sub-parts (e.g.~an eye contains an iris and eyelashes, a
keypad is composed of buttons). Such a hierarchical physical structure
certainly imposes strong constraints on the spatial disposition of the
visual features.

Our third contribution is to show how highly informative spatial
combinations of visual features can be iteratively constructed in the
framework of RLVC. This result is promising for it permits the
construction of features at increasingly higher levels of
discriminative power, enabling us to tackle visual tasks that are
unsolvable using individual point features alone. To the best of our
knowledge, this extension to RLVC appears to be the very first attempt
to build visual feature hierarchies in a closed-loop, interactive and
purposive learning process.

%% file: reinforcement-learning.tex
\section{An Overview of Reinforcement Learning}
\label{def-rl}

Our framework relies on the theory of RL, which is introduced in this
section.  In RL, the environment is traditionally modeled as a {\it
Markov Decision Process\/} (MDP). An MDP is a tuple $\langle S,A,{\cal
T},{\cal R}\rangle$, where $S$ is a finite set of states, $A$ is a
finite set of {\it actions\/}, $\cal T$ is a probabilistic {\it
transition function\/} from $S\times A$ to $S$, and $\cal R$ is a {\it
reinforcement function\/} from $S\times A$ to $\Rset$. An MDP obeys
the following discrete-time dynamics: If at time $t$, the agent takes
the action $a_t$ while the environment lies in a state $s_t$, the
agent perceives a numerical reinforcement $r_{t+1}={\cal R}(s_t,a_t)$,
then reaches some state $s_{t+1}$ with probability ${\cal
T}(s_t,a_t,s_{t+1})$. Thus, from the point of view of the agent, an
{\it interaction\/} with the environment is defined as a quadruple
$\langle s_t,a_t,r_{t+1},s_{t+1}\rangle$. Note that the definition of
Markov decision processes assumes the {\it full observability\/} of
the state space, which means that the agent is able to distinguish
between the states of the environment using only its sensors. This
allows us to talk indifferently about states and percepts. In visual
tasks, $S$ is a set of images.

A {\it percept-to-action mapping\/} is a fixed probabilistic function
$\pi:S\mapsto A$ from states to actions. A percept-to-action mapping
tells the agent the probability with which it should choose an action
when faced with some percept. In RL terminology, such a mapping is
called a {\it stationary Markovian control policy\/}.  For an infinite
sequence of interactions starting in a state $s_t$, the {\it
discounted return\/} is
\begin{equation}
R_t=\sum_{i=0}^\infty \gamma^i r_{t+i+1},
\end{equation}
where $\gamma\in [0,1[$ is the {\it discount factor\/} that gives the
current value of the future reinforcements.  The {\it Markov decision
problem\/} for a given MDP is to find an optimal percept-to-action
mapping that maximizes the expected discounted return, whatever the
starting state is. It is possible to prove that this problem is
well-defined, in that such an optimal percept-to-action mapping always
exists~\cite{Bellman:book-1957}.

Markov decision problems can be solved using {\it Dynamic
Programming\/} (DP)
algorithms \cite{Howard:book-1960,Derman:book-1970}. Let $\pi$ be a
percept-to-action mapping. Let us call the {\it state-action value
function\/} $Q^\pi(s,a)$ of $\pi$, the function giving for each state
$s\in S$ and each action $a\in A$ the expected discounted return
obtained by starting from state $s$, taking action $a$, and thereafter
following the mapping $\pi$:
\begin{equation}
Q^\pi(s,a)={\rm E}^\pi\left\{R_t\mid s_t=s, a_t=a\right\},
\end{equation}
where ${\rm E}^\pi$ denotes the expected value if the agent follows
the mapping $\pi$.  Let us also define the {\it $H$ transform\/} from
$Q$ functions to $Q$ functions as
\begin{equation}
\label{h-transform}
(HQ)(s,a)={\cal R}(s,a) + \gamma \sum_{s'\in S}{\cal T}(s,a,s') 
\max_{a'\in A}Q(s',a'),
\end{equation}
for all $s\in S$ and $a\in A$. Note that the $H$ transform is equally
referred to as the {\it Bellman backup operator\/} for state-action
value functions.  All the optimal mappings for a given MDP share the
same $Q$ function, denoted $Q^*$ and called the {\it optimal
state-action value function\/}, that always exists and that satisfies
Bellman's so-called optimality equation~\cite{Bellman:book-1957}
\begin{equation}
\label{bellman}
HQ^*=Q^*.
\end{equation}
Once the optimal state-action value function $Q^*$ is known, an
optimal deterministic percept-to-action mapping $\pi^*$ is easily
derived by choosing
\begin{equation}
\pi^*(s)=\argmax_{a\in A} Q^*(s,a),
\end{equation}
for each $s\in S$. Another very useful concept from the DP theory is
that of {\it optimal value function\/} $V^*$. For each state $s\in S$,
$V^*(s)$ corresponds to the expected discounted return when the agent
always chooses the optimal action in each encountered state, i.e.
\begin{equation}
V^*(s) = \max_{a\in A} Q^*(s,a).
\end{equation}

Dynamic Programming includes the well-known {\it Value
Iteration\/}~\cite{Bellman:book-1957}, {\it Policy
Iteration\/}~\cite{Howard:book-1960} and {\it Modified Policy
Iteration\/}~\cite{Puterman:MS1978} algorithms. Value Iteration learns
the optimal state-action value function $Q^*$, whereas Policy
Iteration and Modified Policy Iteration directly learn an optimal
percept-to-action mapping.

RL is a set of algorithmic methods for solving Markov decision
problems when the underlying MDP is not known
\cite{Bertsekas:book-1996,Kaelbling:JAIR1996,Sutton:book-1998}. Precisely,
RL algorithms do not assume the knowledge of $\cal T$ and $\cal
R$. The input of RL algorithms is basically a sequence of interactions
$\langle s_t,a_t,r_{t+1},s_{t+1}\rangle$ of the agent with its
environment. RL techniques are often divided in two categories:
\begin{enumerate}
\item {\it Model-based methods\/} that first build an estimate of the
underlying MDP (e.g.~by computing the relative frequencies that
appear in the sequence of interactions), and then use classical DP
algorithms such as Value or Policy Iteration;
\item {\it Model-free methods\/} such as {\it
SARSA\/}~\cite{Rummery:tech-94},
$TD(\lambda)$~\cite{Barto:TSMC1983,Sutton:ML1988}, and the popular
{\it $Q$-learning\/}~\cite{Watkins:phd-1989}, that do not compute such
an estimate.
\end{enumerate}

%% file: rlvc.tex
\section{Reinforcement Learning of Visual Classes}

As discussed in the Introduction, we propose to insert an image
classifier before the RL algorithm. This classifier maps the visual
stimuli to a set of visual classes according to the local-appearance
paradigm, by focusing the attention of the agent on highly distinctive
local descriptors detected at the interest points of the images.

\subsection{Incremental Discretization of the Visual Space}

Formally, let us call $D$, the infinite set of local descriptors that
can be spanned through the chosen local description method. The
elements of $D$ will be equivalently referred to as {\it visual
features\/}. Usually, $D$ corresponds to $\Rset^n$ for some
$n\geq 1$. We assume the existence of a {\it visual feature
detector\/}, that is a Boolean function ${\cal D}:S\times
D\mapsto{\cal B}$ testing whether a given image exhibits a given local
descriptor at one of its interest points~\cite{Schmid:IJCV2000}. Any
suitable metric can be used to test the similarity of two visual
features, e.g.~Mahalanobis or Euclidean distance.

The image classifier is iteratively refined.  Because of this
incremental process, a natural way to implement the image classifiers
is to use binary decision trees. Each of their internal nodes is
labeled by the visual feature, the presence of which is to be tested
in that node. The leaves of the trees define a set of visual classes,
which is hopefully much smaller than the original visual space, and
upon which it is possible to apply directly any usual RL algorithm. To
classify an image, the system starts at the root node, then progresses
down the tree according to the result of the feature detector $\cal D$
for each visual feature found during the descent, until reaching a
leaf.

To summarize, RLVC builds a sequence ${\cal C}_0,{\cal C}_1,{\cal
C}_2,\ldots$ of growing decision trees, in a sequence of attempts to
remove perceptual aliasing. The initial classifier ${\cal C}_0$ maps
all of its input images in a single visual class $V_{0,1}$. At any stage
$k$, the classifier ${\cal C}_k$ partitions the visual perceptual
space $S$ into a finite number $m_k$ of visual classes
$\{V_{k,1},\ldots,V_{k,m_k}\}$.

\subsection{Learning Architecture}

The resulting learning architecture has been called {\it Reinforcement
Learning of Visual Classes\/} (RLVC) \cite{Jodogne:ICML2005}. The
basic idea behind our algorithms, namely the iterative learning of a
decision tree, is primarily motivated by adaptive-resolution techniques
that have been previously introduced in reinforcement learning, and
notably by McCallum's {\it U-Tree\/}
algorithm~\cite{McCallum:phd-1996}. In this section, this idea is
showed to be extremely fruitful when suitably adapted to visual
spaces. The links between RLVC and adaptive-resolution techniques will
be more thoroughly discussed in Section~\ref{sec:related-work}.

\begin{figure*}[t]
\begin{center}
\includegraphics[width=13cm]{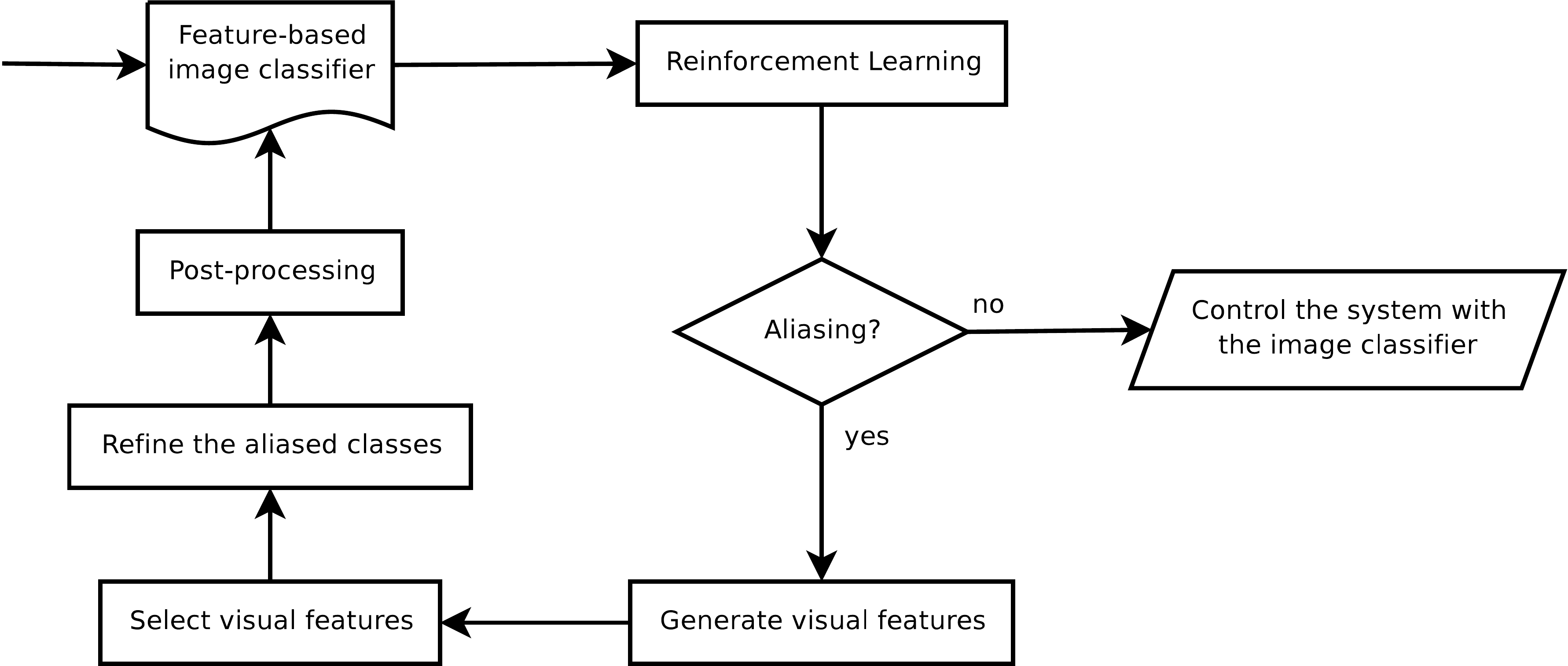}
\end{center}
\caption{The different components of the RLVC
algorithm.\label{fig-global}}
\end{figure*}

The components of RLVC are depicted in Figure~\ref{fig-global}. An
in-depth discussion of each of those components will be given in the
next sections. For the time being, we review each of them:
\begin{description}
\item[RL algorithm:] For each classifier in the sequence, an
arbitrary, standard RL algorithm is applied. This provides
information such as the optimal state-action function, the optimal
value function or the optimal policy that are induced by the current
classifier ${\cal C}_k$. For the purpose of these computations, either
new interactions can be acquired, or a database of previously
collected interactions can be exploited. This component is covered in
Sections~\ref{mdp-projection} and~\ref{mdp-optimal}.

\item[Aliasing detector:] Until the agent has learned the visual
classes required to complete its task, from the viewpoint of the
embedded RL algorithm, the input space is only partially
observable. The aliasing detector extracts the classes in which
perceptual aliasing occurs, through an analysis of the Bellman
residuals. Indeed, as explained in Section~\ref{aliasing-measure},
there exist tight relations between perceptual aliasing and Bellman
residuals.  If no aliased class is detected, RLVC stops.

\item[Feature generator:] After having applied the RL algorithm, a
database of interactions $\langle s_t, a_t, r_{t+1}, s_{t+1}\rangle$
is available. The feature generator component produces a set $F$
of candidate visual features for each aliased class $V_{k,i}$. The
features that are used to refine a classifier will be chosen among
this set of candidates. This step is further exposed in
Sections~\ref{feature-selection} and~\ref{sec-spatial}.

\item[Feature selector:] Once the set of candidate features $F$ is
built for the aliased visual class $V_{k,i}$, this component selects the
visual feature $f^*\in F$ that best reduces the perceptual
aliasing. If no candidate feature is discriminant, the component
returns the conventional bottom symbol $\perp$. The feature selector
of RLVC is described in Section~\ref{feature-selection}.

\item[Classifier refinement:] The leaves that correspond to the
aliased classes in the feature-based image classifier are replaced by
an internal node testing the presence or absence of the selected
visual features.

\item[Post-processing:] This optional component is invoked after every
refinement, and corresponds to techniques for fighting
overfitting. Details are given in Section~\ref{sec-compact}.
\end{description}

\begin{algorithm}[t]
\caption{--- General structure of RLVC\label{algo-rlvc}}
\begin{algorithmic}[1]
 \STATE $k \leftarrow 0$
 \STATE $m_k\leftarrow 1$
 \STATE ${\cal C}_k \leftarrow$ binary decision tree with one leaf
 \REPEAT
  \STATE Collect $N$ interactions $\langle
  s_t,a_t,r_{t+1},s_{t+1}\rangle$
  \STATE Apply an arbitrary RL algorithm on the sequence
  that is mapped through ${\cal C}_k$
  \STATE ${\cal C}_{k+1} \leftarrow {\cal C}_k$
  \FORALL{$i \in \{1,\ldots,m_k\}$}
   \IF{$\mbox{\tt aliased}(V_{k,i})$}
    \STATE $F\leftarrow \mbox{\tt generator}
    \left(\left\{s_t\mid {\cal C}_k(s_t)=V_{k,i}\right\}\right)$
    \STATE $f^*\leftarrow \mbox{\tt selector}\left(V_{k,i}, F\right)$
    \IF{$f^*\not=\ \perp$}
    \STATE In ${\cal C}_{k+1}$, refine $V_{k,i}$ by testing $f^*$
    \STATE $m_{k+1}\leftarrow m_{k+1}+1$
    \ENDIF
   \ENDIF
  \ENDFOR
  \STATE $k\leftarrow k+1$
  \STATE $\mbox{\tt post-process}({\cal C}_k)$
 \UNTIL ${\cal C}_k = {\cal C}_{k-1}$
\end{algorithmic}
\end{algorithm}

The general outline of RLVC is described in
Algorithm~\ref{algo-rlvc}. Note that in all the experiments that are
contained in this paper, model-based RL algorithms were applied to
static databases of interactions at the fifth step of
Algorithm~\ref{algo-rlvc}. These databases were collected using a
fully randomized exploration policy. This choice was only guided by
the ease of implementation and presentation; any other way of
collecting experience could be used as well, for example by
re-sampling a new database of interactions at each iteration of
RLVC. The crucial point here is that RLVC generates a representation
for visual control policies only from a set of collected visuomotor
experience, which makes RLVC interactive. The following sections
describe the remaining algorithms, namely {\tt aliased}, {\tt
generator}, {\tt selector} and {\tt post-process}.

\subsection{Detection of the Aliased Visual Classes}
\label{aliasing-criterion}

We now discuss how aliasing can be detected in a classifier ${\cal
C}_k$.

\subsubsection{Projection of an MDP through an Image Classifier}
\label{mdp-projection}

Formally, any image classifier ${\cal C}_k$ converts a sequence of $N$
interactions
\begin{equation*}
\langle s_t, a_t, r_{t+1}, s_{t+1}\rangle,
\end{equation*}
to a {\it mapped sequence\/} of $N$ quadruples
\begin{equation*}
\langle {\cal C}_k(s_t), a_t, r_{t+1}, {\cal C}_k(s_{t+1}) \rangle,
\end{equation*}
upon which the embedded RL algorithm is applied. Let us define the
{\it mapped MDP\/} ${\cal M}_k$ as the MDP
\begin{equation*}
\langle S_k, A, {\cal T}_k, {\cal R}_k \rangle,
\end{equation*}
that is obtained from the mapped sequence, where $S_k$ is the set of
visual classes that are known to ${\cal C}_k$, and where ${\cal T}_k$
and ${\cal R}_k$ have been computed using the relative frequencies in
the mapped sequence, as follows.

Consider two visual classes $V,V'\in \{V_{k,1},\ldots,V_{k,m_k}\}$ and
one action $a\in A$. We define the following functions:
\begin{itemize}
\item $\delta_t(V,a)$ equals $1$ if ${\cal C}_k(s_t)=V$ and $a_t=a$,
and $0$ otherwise;
\item $\delta_t\left(V,a,V'\right)$ equals $1$ if ${\cal C}_k(s_t)=V$,
${\cal C}_k(s_{t+1})=V'$ and $a_t=a$, and $0$ otherwise;
\item $\eta(V,a)$ is the number of $t$'s such that $\delta_t(V,a)=1$.
\end{itemize}
Using this notation, we can write:
\begin{itemize}
\item $S_k=\left\{V_{k,1},\ldots,V_{k,m_k}\right\}$;
\item ${\cal T}_k \left(V,a,V'\right) = 
\sum_{t=1}^N \delta_t\left(V,a,V'\right) / \eta(V,a)$;
\item ${\cal R}_k(V,a) = \sum_{t=1}^N r_t \delta_t(V,a) / \eta(V,a)$.
\end{itemize}

\subsubsection{Optimal $Q$ Function for a Mapped MDP}
\label{mdp-optimal}

Each mapped MDP ${\cal M}_k$ induces an optimal $Q$ function on the
domain $S_k\times A$ that will be denoted $Q'^*_k$.  Computing
$Q'^*_k$ can be difficult: In general, there may exist no MDP defined
on the state space $S_k$ and on the action space $A$ that can generate
a given mapped sequence, since the latter is not necessarily Markovian
anymore. Thus, if some RL algorithm is run on the mapped sequence, it
might not converge toward $Q'^*_k$, or not even converge at
all. However, when applied on a mapped sequence, any model-based RL
method (cf. Section~\ref{def-rl}) can be used to compute $Q'^*_k$ if
${\cal M}_k$ is used as the underlying model. Under some conditions,
$Q$-learning also converges to the optimal $Q$ function of the mapped
MDP~\cite{Singh:NIPS1995}.

In turn, the function $Q'^*_k$ induces another $Q$ function on the
initial domain $S\times A$ through the relation:
\begin{equation}
Q^*_k(s,a)=Q'^*_k\left({\cal C}_k(s),a\right),
\end{equation}
In the absence of aliasing, the agent would perform optimally, and
$Q^*_k$ would correspond to $Q^*$, according to Bellman theorem that
states the uniqueness of the optimal $Q$ function
(cf. Section~\ref{def-rl}). By Equation~\ref{bellman}, the function
\begin{equation}
\label{eqn:bellman-residual}
B_k(s,a) = (HQ^*_k)(s,a)-Q^*_k(s,a)
\end{equation}
is therefore a measure of the aliasing induced by the image classifier
${\cal C}_k$. In RL terminology, $B_k$ is {\it Bellman residual\/} of
the function $Q^*_k$~\cite{Sutton:ML1988}. The basic idea behind RLVC
is to refine the states that have a non-zero Bellman residual.

\subsubsection{Measuring Aliasing}
\label{aliasing-measure}

Consider a time stamp $t$ in a database of interactions $\langle
s_t,a_t,r_{t+1},s_{t+1}\rangle$. According to
Equation~\ref{eqn:bellman-residual}, the Bellman residual that
corresponds to the state-action pair $(s_t,a_t)$ equals
\begin{equation}
\label{eqn:bellman-residual-2}
B_k(s_t,a_t) = {\cal R}(s_t,a_t) + \gamma \sum_{s'\in S}{\cal
T}(s_t,a_t,s') \max_{a'\in A}Q^*_k(s',a') - Q^*_k(s_t,a_t).
\end{equation}
Unfortunately, the RL agent does not have access to the transition
probabilities $\cal T$ and to the reinforcement function $\cal R$ of
the MDP modeling the environment. Therefore,
Equation~\ref{eqn:bellman-residual-2} cannot be directly evaluated.
A similar problem arises in the $Q$-learning~\cite{Watkins:phd-1989}
and the Fitted $Q$ Iteration~\cite{Ernst:JMLR2005} algorithms. These
algorithms solve this problem by considering the stochastic version of
the time difference that is described by
Equation~\ref{eqn:bellman-residual-2}: The value
\begin{equation}
\sum_{s'\in S} {\cal T}(s_t,a_t,s') \max_{a'\in A} Q^*_k(s',a')
\end{equation}
can indeed be estimated as
\begin{equation}
\max_{a'\in A} Q^*_k(s',a'),
\end{equation}
if the successor $s'$ is chosen with probability ${\cal
T}(s_t,a_t,s')$. But following the transitions of the environment
ensures making a transition from $s_t$ to $s_{t+1}$ with probability
${\cal T}(s_t,a_t,s_{t+1})$. Thus
\begin{eqnarray}
\Delta_t & = & r_{t+1} + \gamma \max_{a'\in A} Q^*_k(s_{t+1}, a') -
Q^*_k(s_t,a_t) \\
\label{eqn-aliasing}
& = & r_{t+1} + \gamma \max_{a'\in A} Q'^*_k\left({\cal C}_k
\left(s_{t+1}\right),a'\right) -
Q'^*_k ({\cal C}_k (s_t),a)
\end{eqnarray}
is an unbiased estimate of the Bellman residual for the state-action
pair $(s_t,a_t)$~\cite{Jaakkola:ANIPS94}.\footnote{It is worth
noticing that $\alpha_t \Delta_t$ corresponds to the updates that
would be applied by $Q$-learning, where $\alpha_t$ is known as the
{\it learning rate\/} at time $t$.} Very importantly, if the system is
deterministic and in the absence of perceptual aliasing, these
estimates are equal to zero. Therefore, a nonzero $\Delta_t$
potentially indicates the presence of perceptual aliasing in the
visual class $V_t={\cal C}_k(s_t)$ with respect to action $a_t$. Our
criterion for detecting the aliased classes consists in computing the
$Q'^*_k$ function, then in sweeping again all the interactions
$\langle s_t,a_t,r_{t+1},s_{t+1} \rangle$ to identify nonzero
$\Delta_t$. In practice, we assert the presence of aliasing if the
variance of the $\Delta_t$ exceeds a given threshold $\tau$. This is
summarized in Algorithm~\ref{algo-aliased}, where $\sigma^2(\cdot)$
denotes the variance of a set of samples.

\begin{algorithm}[t]
\caption{--- Aliasing Criterion\label{algo-aliased}}
\begin{algorithmic}[1]
 \NAME{$\mbox{\tt aliased}(V_{k,i})$}
 \FORALL{$a\in A$}
  \STATE $\Delta \leftarrow \{\Delta_t \mid {\cal C}_k(s_t)=V_{k,i} 
    \wedge a_t=a\}$
  \IF{$\sigma^2(\Delta) > \tau$}
    \STATE {\bf return} true
  \ENDIF
 \ENDFOR
 \STATE {\bf return} false
 \ENDNAME
\end{algorithmic}
\end{algorithm}

\subsection{Generation and Selection of Distinctive Visual Features}
\label{feature-selection}

Once aliasing has been detected in some visual class $V_{k,i}\in S_k$
with respect to an action $a$, we need to select a local descriptor
that best explains the variations in the set of $\Delta_t$ values
corresponding to $V_{k,i}$ and $a$. This local descriptor is to be
chosen among a set of candidate visual features $F$. 

\subsubsection{Extraction of Candidate Features}

Informally, the canonical way of building $F$ for a visual class
$V_{k,i}$ consists in:
\begin{enumerate}
\item identifying all collected visual percepts $s_t$ such that ${\cal
C}_k(s_t)=V_{k,i}$,
\item locating all the interest points in all the selected images
$s_t$, then
\item adding to $F$ the local descriptor of all those interest
points.
\end{enumerate}
The corresponding feature generator is detailed in
Algorithm~\ref{algo-simple-generator}. In the latter algorithm,
$\mbox{\tt descriptor}(s,x,y)$ returns the local description of the
point at location $(x,y)$ in the image $s$, and $\mbox{\tt
symbol}({\bf d})$ returns the symbol that corresponds to the local
descriptor ${\bf d}\in F$ according to the used metric. However, more
complex strategies for generating the visual features can be
used. Such a strategy that builds spatial combinations of individual
point features will be presented in Section~\ref{sec-spatial}.

\begin{algorithm}[t]
\caption{--- Canonical feature generator \label{algo-simple-generator}}
\begin{algorithmic}[1]
  \NAME{$\mbox{\tt generator}(\{s_1,\ldots,s_n\})$}
  \STATE $F\leftarrow \{\}$
  \FORALL{$i \in \{1,\ldots,n\}$}
  \FORALL{$(x,y)$ such that $(x,y)$ is an interest point of $s_i$}
  \STATE $F\leftarrow F\cup \{\mbox{\tt symbol}
  (\mbox{\tt descriptor}(s_i,x,y))\}$
  \ENDFOR
  \ENDFOR
  \STATE {\bf return} $F$
  \ENDNAME
\end{algorithmic}
\end{algorithm}

\subsubsection{Selection of Candidate Features}

The problem of choosing the candidate feature that most reduces the
variations in a set of real-valued Bellman residuals is a regression
problem, for which we suggest an adaptation of a popular splitting
rule used in the CART algorithm for building regression
trees~\cite{Breiman:book-1984}.\footnote{Note that in our previous
work, we used a splitting rule that is borrowed from the building of
classification trees~\cite{Quinlan:book-1993,Jodogne:ICML2005}.}

In CART, variance is used as an impurity indicator: The split that is
selected to refine a particular node is the one that leads to the
greatest reduction in the sum of the squared differences between the
response values for the learning samples corresponding to the node and
their mean. More formally, let $S=\{ \langle {\bf x}_i, y_i\rangle\}$
be a set of learning samples, where ${\bf x}_i\in \Rset^n$ are input
vectors of real numbers, and where $y_i\in \Rset$ are real-valued
outputs. CART selects the following candidate feature:
\begin{equation}
f^* = \argmin_{v\in F} \left(p^v_\oplus\cdot
\sigma^2\left(S^v_\oplus\right) + p^v_\ominus\cdot
\sigma^2\left(S^v_\ominus\right)\right),
\end{equation}
where $p^v_\oplus$ (resp. $p^v_\ominus$) is the proportion of samples
that exhibit (resp. do not exhibit) the feature $v$, and where
$S^v_\oplus$ (resp. $S^v_\ominus$) is the set of samples that exhibit
(resp. do not exhibit) the feature $v$. This idea can be directly
transferred in our framework, if the set of ${\bf x}_i$ corresponds to
the set of interactions $\langle s_t, a_t, r_{t+1}, s_{t+1}\rangle$,
and if the set of $y_i$ corresponds to the set of $\Delta_t$. This is
written explicitly in Algorithm~\ref{algo-selection}.

\begin{algorithm}[t]
\caption{--- Feature Selection\label{algo-selection}}
\begin{algorithmic}[1]
 \NAME{$\mbox{\tt selector}(V_{k,i}, F)$}
 \STATE $f^*\leftarrow \ \perp$~~~~~~~~\COMMENT{Best feature found so far}
 \STATE $r^*\leftarrow +\infty$~~~~~\COMMENT{Variance reduction induced by $f^*$}
 \FORALL{$a\in A$}
 \STATE $T \leftarrow \{ t \mid {\cal C}_k(s_t)=V_{k,i}
  \mbox{~and~}a_t=a\}$
 \FORALL{visual feature $f\in F$}
   \STATE $S_\oplus \leftarrow \{ \Delta_t \mid t \in T \mbox{~and~}
   s_t \mbox{~exhibits~} f\}$
   \STATE $S_\ominus \leftarrow \{ \Delta_t \mid t \in T \mbox{~and~}
   s_t \mbox{~does not exhibit~} f\}$
   \STATE $s_\oplus \leftarrow |S_\oplus| / |T|$
   \STATE $s_\ominus \leftarrow |S_\ominus| / |T|$
   \STATE $r\leftarrow s_\oplus\cdot
   \sigma^2\left(S_\oplus\right) + s_\ominus\cdot
   \sigma^2\left(S_\ominus\right)$
   \IF{$r < r^*$ {\bf and} the distributions $(S_\oplus,S_\ominus)$ are
   significantly different}
    \STATE $f^*\leftarrow f$
    \STATE $r^*\leftarrow s$
   \ENDIF
  \ENDFOR
 \ENDFOR
 \STATE {\bf return} $f^*$
 \ENDNAME
\end{algorithmic}
\end{algorithm}

Our algorithms exploit the stochastic version of Bellman residuals.
Of course, real environments are in general non-deterministic, which
generates variations in Bellman residuals that are not a consequence
of perceptual aliasing. RLVC can be made somewhat robust to such a
variability by introducing a statistical hypothesis test: For each
candidate feature, a Student's $t-$test is used to decide whether the
two sub-distributions the feature induces are significantly different.
This approach is also used in U-Tree~\cite{McCallum:phd-1996}.

\subsection{Illustration on a Simple Navigation Task}

We have evaluated our system on an abstract task that closely
parallels a real-world scenario while avoiding any unnecessary
complexity. As a consequence, the sensor model we use may seem
unrealistic; a better visual sensor model will be exploited in
Section~\ref{sec:realistic}.

\begin{figure}[t]
\centerline{\includegraphics[width=10cm]{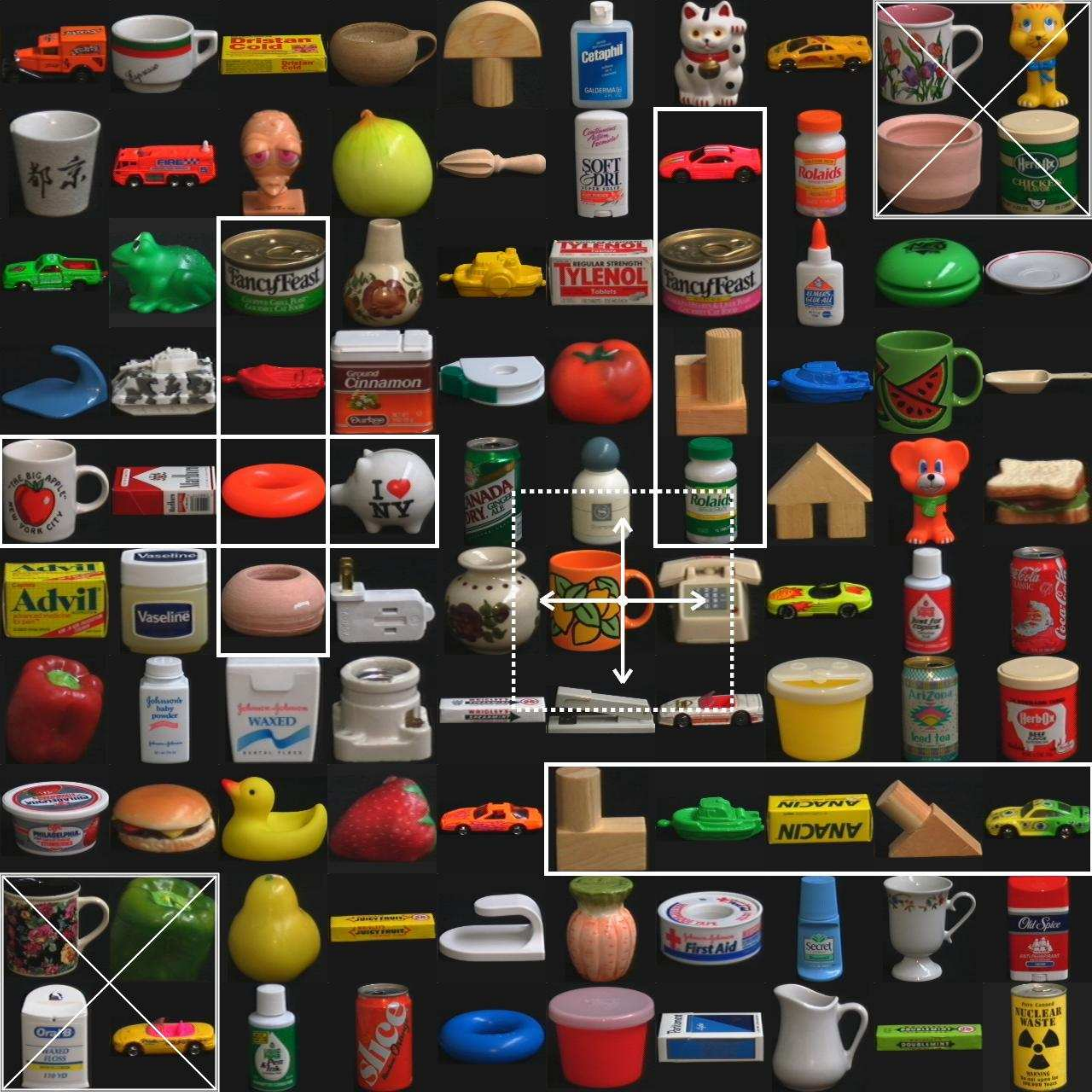}}
\caption{A continuous, noisy navigation task. The exits of the maze
are indicated by boxes with a cross. Walls of glass are identified by
solid lines. The agent is depicted at the center of the figure.  Each
one of the four possible moves is represented by an arrow, the length
of which corresponds to the resulting move.  The sensors return a
picture that corresponds to the dashed portion of the
image.\label{fig-navigation}}
\end{figure}

RLVC has succeeded at solving the continuous, noisy visual navigation
task depicted in Figure~\ref{fig-navigation}. The goal of the agent is
to reach as fast as possible one of the two exits of the maze. The set
of possible locations is continuous. At each location, the agent has
four possible actions: Go up, right, down, or left. Every move is
altered by a Gaussian noise, the standard deviation of which is $2\%$
the size of the maze. Glass walls are present in the maze.  Whenever a
move would take the agent into a wall or outside the maze, its
location is not changed.

\begin{figure}[t]
\centerline{\includegraphics[width=12cm]{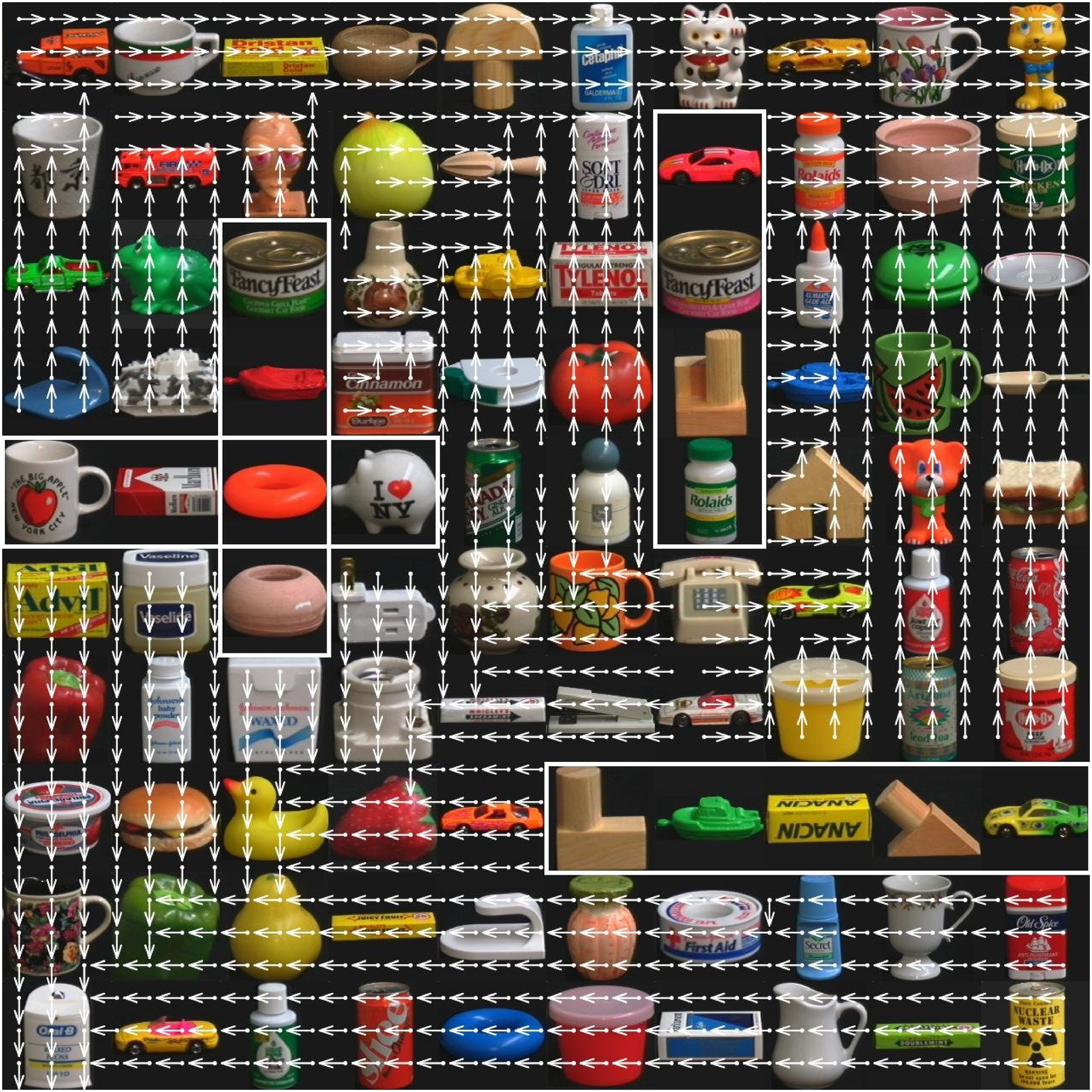}}
\caption{The deterministic image-to-action mapping that results from
RLVC, sampled at regularly-spaced points. It manages to choose the
correct action at each location.\label{fig-result}}
\end{figure}

The agent earns a reward of 100 when an exit is reached. Any other
move, including the forbidden ones, generates zero reinforcement. When
the agent succeeds in escaping the maze, it arrives in a terminal
state in which every move gives rise to a zero reinforcement. In this
task, $\gamma$ was set to $0.9$.  Note that the agent is faced with
the delayed reward problem, and that it must take the distance to the
two exits into consideration for choosing the most attractive one.

The maze has a ground carpeted with a color image of $1280\times 1280$
pixels that is a montage of pictures from the COIL-100
database~\cite{COIL100-96}.  The agent does not have direct access to
its $(x,y)$ position in the maze. Rather, its sensors take a picture
of a surrounding portion of the ground. This portion is larger than
the blank areas, which makes the input space fully
observable. Importantly, the glass walls are transparent, so that the
sensors also return the portions of the tapestry that are behind
them. Therefore, there is no way for the agent to directly locate the
walls. It is obliged to identify them as the regions of the maze in
which an action does not change its location.

In this experiment, we have used color differential invariants as
visual features~\cite{Gouet:CBAIVL2001}. The entire tapestry includes
2298 different visual features. RLVC selected 200 features,
corresponding to a ratio of $9\%$ of the entire set of possible
features. The computation stopped after the generation of 84 image
classifiers (i.e.~when $k$ reached 84), which took 35 minutes on a
2.4GHz Pentium IV using databases of 10,000 interactions. 205 visual
classes were identified. This is a small number, compared to the
number of perceptual classes that would be generated by a
discretization of the maze when the agent knows its $(x,y)$
position. For example, a reasonably sized $20\times 20$ grid leads to
400 perceptual classes.

\begin{figure}
\noindent
\centerline{\begin{tabular}{cc}
\includegraphics[width=6cm]{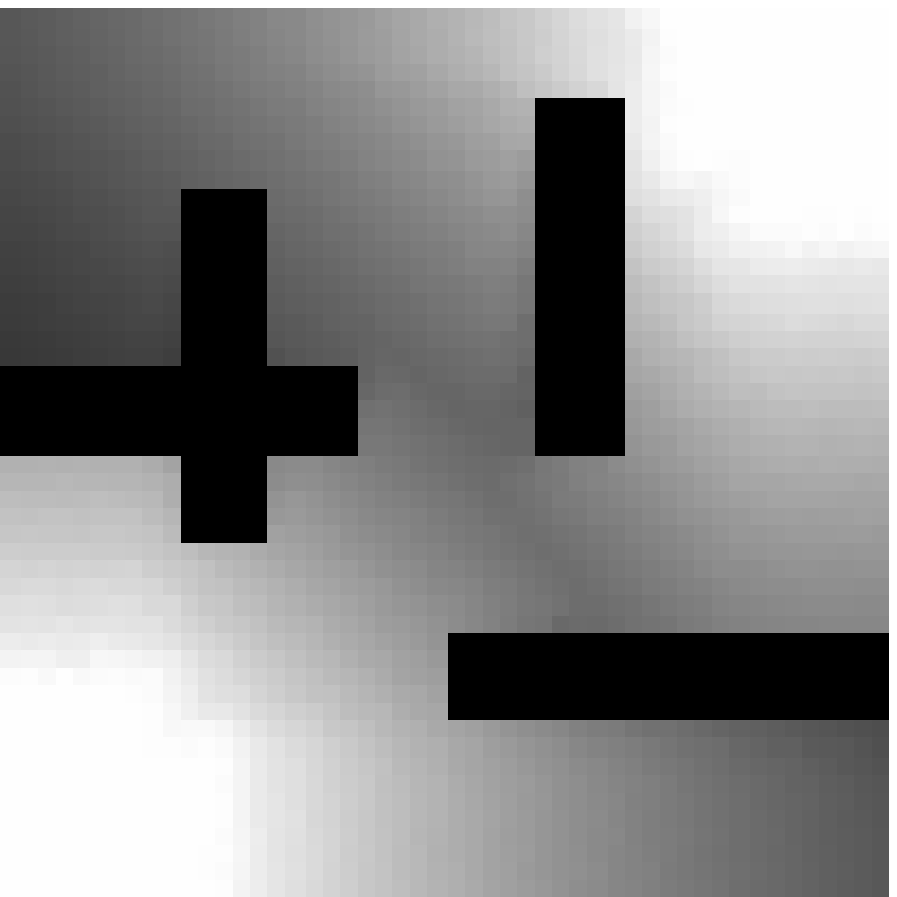} &
\includegraphics[width=6cm]{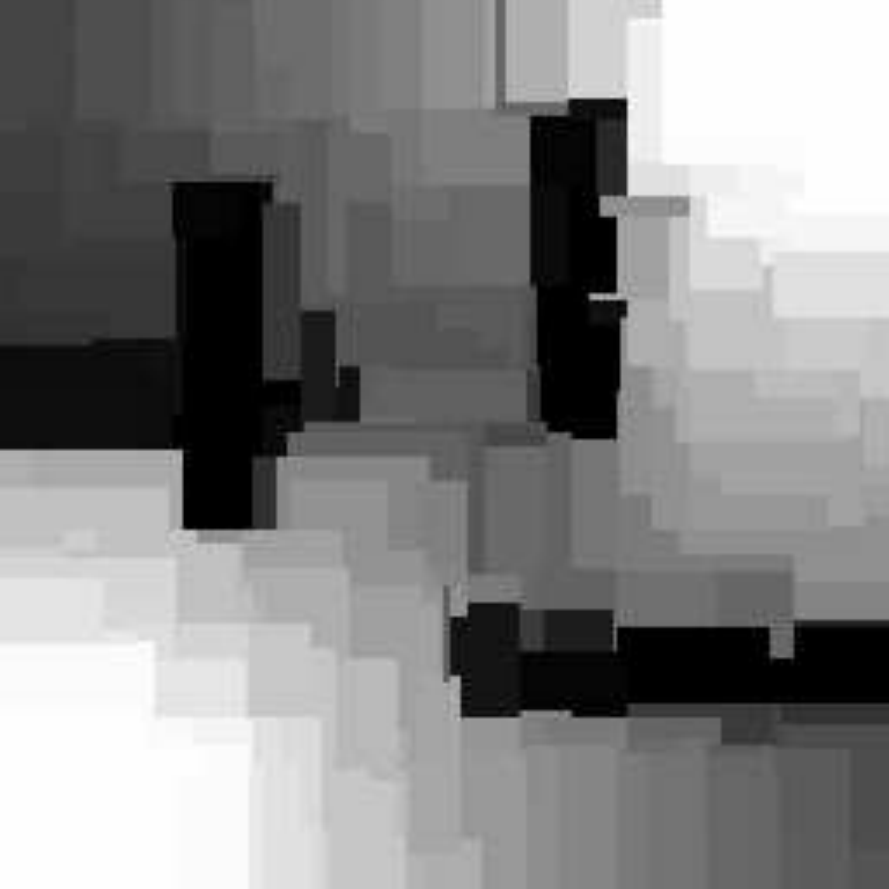}  \\[.5em]
(a) & (b)
\end{tabular}}
\caption{(a) The optimal value function, when the agent has direct
access to its $(x,y)$ position in the maze and when the set of
possible locations is discretized into a $50\times 50$ grid. The
brighter the location, the greater its value.  (b) The final value
function obtained by RLVC.\label{fig-values}}
\end{figure}

Figure~\ref{fig-result} shows the optimal, deterministic
image-to-action mapping that results from the last obtained image
classifier ${\cal C}_k$:
\begin{equation}
\pi^*(s)=\argmax_{a\in A} Q^*_k(s,a)=Q'^*_k\left({\cal C}_k(s),a\right).
\end{equation}
Figure~\ref{fig-values} compares the optimal value function of the
discretized problem with the one obtained through RLVC. The similarity
between the two pictures indicates the soundness of our
approach. Importantly, RLVC operates with neither pretreatment, nor
human intervention. The agent is initially not aware of which visual
features are important for its task. Moreover, the interest of
selecting descriptors is clear in this application: A direct, tabular
representation of the $Q$ function considering all the Boolean
combinations of features would have $2^{2298}\times 4$ cells.

\begin{figure}[p]
\begin{center}
\begin{tabular}{ccc}
\includegraphics[angle=90,height=18.6cm]{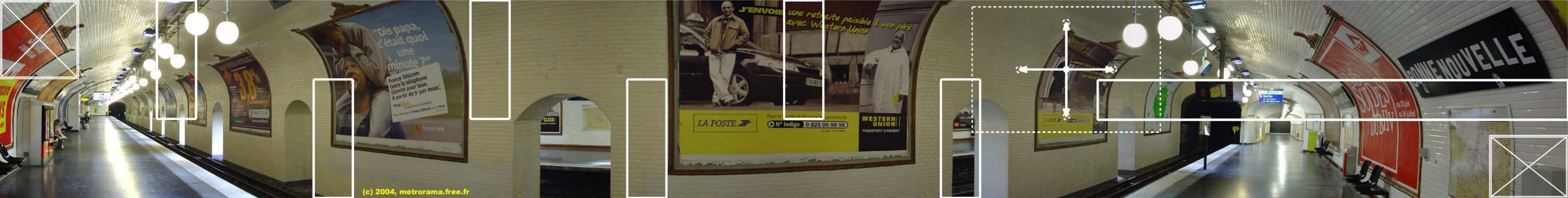} &&
\includegraphics[angle=90,height=18.6cm]{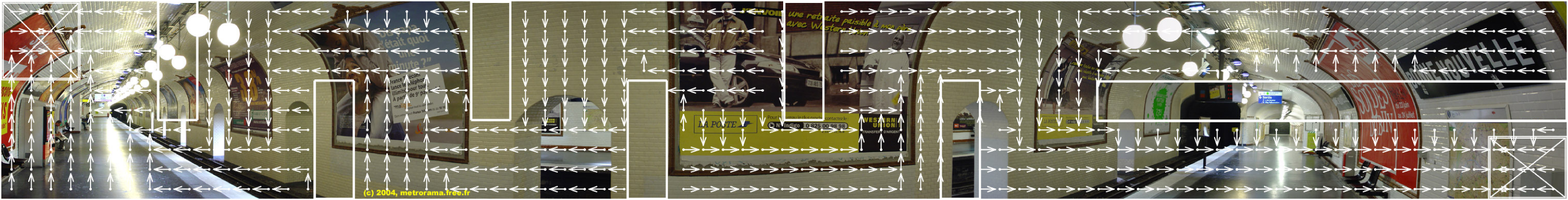} \\[.5em]
(a) && (b)
\end{tabular}
\end{center}
\caption{(a) A navigation task with a real-world image, using the same
conventions than Figure~\ref{fig-navigation}. (b) The deterministic
image-to-action mapping computed by RLVC. \label{fig-subway}}
\end{figure}

The behavior of RLVC on real-word images has also been investigated.
The navigation rules were kept identical, but the tapestry was
replaced by a panoramic photograph of $3041\times 384$ pixels of a
subway station, as depicted in Figure~\ref{fig-subway}. RLVC took 101
iterations to compute the mapping at the right of
Figure~\ref{fig-subway}. The computation time was 159 minutes on a
2.4GHz Pentium IV using databases of 10,000 interactions. 144 distinct
visual features were selected among a set of 3739 possible ones,
generating a set of 149 visual classes.  Here again, the resulting
classifier is fine enough to obtain a nearly optimal image-to-action
mapping for the task.

\subsection{Related Work}
\label{sec:related-work}

RLVC can be thought of as performing adaptive discretization of the
visual space on the basis of the presence of visual features. Previous
reinforcement learning algorithms that exploit the presence of
perceptual features in various contexts are now discussed.

\subsubsection{Perceptual Aliasing}

As explained above, the incremental selection of a set of informative
visual features necessarily leads to temporary perceptual aliasing,
which RLVC tries to remove. More generally, perceptual aliasing occurs
whenever an agent cannot always take the right on the basis of its
percepts.

Early work in reinforcement learning has tackled this general problem
in two distinct ways: Either the agent identifies and then avoids
states where perceptual aliasing occurs \cite<as in the {\sc Lion}
algorithm, see>{Whitehead:ML-1991}, or it tries to build a short-term
memory that will allow it to remove the ambiguities on its percepts
\cite<as in the {\it predictive distinctions\/} approach,
see>{Chrisman:NCAI-1992}. Very sketchily, these two algorithms detect
the presence of perceptual aliasing through an analysis of the sign of
$Q$-learning updates. The possibility of managing a short-term memory
has led to the development of the {\it Partially Observable Markov
Decision Processes\/} (POMDP) theory~\cite{Kaelbling:AI1998}, in which
the current state is a random variable of the percepts.

Although these approaches are closely related to the perceptual 
aliasing RLVC temporarily introduces, they do not consider the 
exploitation of perceptual features. Indeed, they tackle a structural 
problem in a given control task, and, as such, they assume that 
perceptual aliasing cannot be removed. As a consequence, these 
approaches are orthogonal to our research interest, since the 
ambiguities RLVC generates can be removed by further refining the 
image classifier. In fact, the techniques above tackle a {\it lack 
of information\/} inherent to the used sensors, whereas our goal is 
to handle a {\it surplus of information\/} related to the high 
redundancy of visual representations.

\subsubsection{Adaptive Resolution in Finite Perceptual Spaces}
\label{sec:adaptive-resolution-finite}

RLVC performs an {\it adaptive\/} discretization of the perceptual
space through an autonomous, task-driven, purposive selection of
visual features. Work in RL that incrementally partitions a large
(either discrete or continuous) perceptual space into a piecewise
constant value function is usually referred to as {\it
adaptive-resolution\/} techniques. Ideally, regions of the perceptual
space with a high granularity should only be present where they are
needed, while a lower resolution should be used elsewhere. RLVC is
such an adaptive-resolution algorithm. We now review several
adaptive-resolution methods that have been previously proposed for
finite perceptual spaces.

The idea of adaptive-resolution techniques in reinforcement learning
goes back to the G~Algorithm~\cite{Chapman:IJACI1991}, and has
inspired the other approaches that are discussed below. The
G~Algorithm considers perceptual spaces that are made up of
fixed-length binary numbers. It learns a decision tree that tests the
presence of informative bits in the percepts. This algorithm uses a
Student's $t$-test to determine if there is some bit $b$ in the
percepts that is mapped to a given leaf, such that the state-action
utilities of states in which $b$ is set are significantly different
from the state-action utilities of states in which $b$ is unset. If
such a bit is found, the corresponding leaf is split. The process is
repeated for each leaf. This method is able to learn compact
representations, even though there is a large number of irrelevant
bits in the percepts. Unfortunately, when a region is split, all the
information associated with that region is lost, which makes for very
slow learning. Concretely, the G~Algorithm can solve a task whose
perceptual space contains $2^{100}$ distinct percepts, which
corresponds to the set of binary numbers with a length of $100$ bits.

McCallum's {\it U-Tree\/} algorithm builds upon this idea by 
combining a ``selective attention'' mechanism inspired by the 
G~Algorithm with a short-term memory that enables the agent to deal 
with partially observable environments~\cite{McCallum:phd-1996}. 
Therefore, McCallum's algorithms are a keystone in reinforcement 
learning, as they unify the G~Algorithm~\cite{Chapman:IJACI1991} with 
Chrisman's predictive distinctions~\cite{Chrisman:NCAI-1992}.

U-Tree incrementally grows a decision tree through Kolmogorov-Smirnov
tests. It has succeeded at learning behaviors in a driving simulator.
In this simulator, a percept consists of a set of 8 discrete variables
whose variation domains contain between 2 and 6 values, leading to a
perceptual space with $2,592$ possible percepts. Thus, the size of the
perceptual space is much smaller than a visual space. However, this
task is difficult because the ``physical'' state space is only
partially observable through the perceptual space: The driving task
contains $21,216$ physical states, which means that several physical
states requiring different reactions can be mapped to the same percept
through the sensors of the agent. U-Tree resolves such ambiguities on
the percepts by testing the presence of perceptual features in the
percepts that have been encountered previously in the history of the
system. To this end, U-Tree manages a short-term memory. In this
paper, partially observable environments are not considered. Our
challenge is rather to deal with huge visual spaces, without
hand-tuned pre-processing, which is in itself a difficult, novel
research direction.

\subsubsection{Adaptive Resolution in Continuous Perceptual Spaces}
\label{sec:adaptive-resolution-continuous}

It is important to notice that all the methods for adaptive resolution
in large-scale, finite perceptual spaces use a fixed set of perceptual
features that is hard-wired. This has to be distinguished from RLVC
that samples visual features from a possibly {\it infinite visual
feature space\/} (e.g.~the set of visual features is infinite), and
that makes no prior assumptions about the maximum number of useful
features. From this point of view, RLVC is closer to
adaptive-resolution techniques for continuous perceptual
spaces. Indeed, these techniques dynamically select new relevant
features from a whole continuum.

The first adaptive-resolution algorithm for continuous perceptual
spaces is the {\sc Darling} algorithm~\cite{Salganicoff:ICML93}. This
algorithm, just like all the current algorithms for continuous
adaptive resolution, splits the perceptual space using thresholds. For
this purpose, {\sc Darling} builds a hybrid decision tree that assigns
a label to each point in the perceptual space. {\sc Darling} is a
fully on-line and incremental algorithm that is equipped with a {\it
forgetting mechanism\/} that deletes outdated interactions. It is
however limited to binary reinforcement signals, and it only takes
immediate reinforcements into account, so that {\sc Darling} is much
closer to supervised learning than to reinforcement learning.

The {\it Parti-Game\/} algorithm~\cite{Moore:ML1995} produces
goal-directed behaviors in continuous perceptual spaces. Parti-Game
also splits regions where it deems it important, using a
game-theoretic approach. Moore and Atkeson show that Parti-Game can
learn competent behavior in a variety of continuous
domains. Unfortunately, the approach is currently limited to
deterministic domains where the agent has a greedy controller and
where the goal state is known. Moreover, this algorithm searches for
{\it any\/} solution to a given task, and does not try to find the
optimal one.

The {\it Continuous U-Tree\/} algorithm is an extension of U-Tree that
is adapted to continuous perceptual spaces~\cite{Uther:AAAI1998}. Just
like {\sc Darling}, Continuous U-Tree incrementally builds a decision
tree that splits the perceptual space into a finite set of hypercubes,
by testing thresholds. Kolmogorov-Smirnov and sum-of-squared-errors
are used to determine when to split a node in the decision
tree. Pyeatt and Howe~\citeyear{Pyeatt:ISAS2001} analyze the
performance of several splitting criteria for a variation of
Continuous U-Tree. They conclude that Student's $t$-test leads to the
best performance, which motivates the use of this statistical test in
RLVC (cf.~Section~\ref{feature-selection}).

Munos and Moore~\citeyear{Munos:ML2002} have proposed {\it Variable
Resolution Grids\/}. Their algorithm assumes that the perceptual space
is a compact subset of Euclidean space, and begins with a coarse,
grid-based discretization of the state space. In contrast with the
other abstract algorithms in this section, the value function and
policy vary linearly within each region. Munos and Moore use Kuhn
triangulation as an efficient way to interpolate the value function
within regions. The algorithm refines its approximation by refining
cells according to a splitting criterion.  Munos and Moore explore
several local heuristic measures of the importance of splitting a cell
including the average of corner-value differences, the variance of
corner-value differences, and policy disagreement. They also explore
global heuristic measures involving the influence and variance of the
approximated system.  Variable Resolution Grids are probably the most
advanced adaptive-resolution algorithm available so far.

\subsubsection{Discussion}

To summarize, several algorithms that are similar in spirit to RLVC
have been proposed over the years. Nevertheless, our work appears to
be the first that can learn direct image-to-action mappings through
reinforcement learning. Indeed, none of the reinforcement learning
methods above combines all the following desirable properties of RLVC:
(1) The set of relevant perceptual features is not chosen a priori by
hand, as the selection process is fully automatic and does not require
any human intervention; (2) visual perceptual spaces are explicitly
considered through appearance-based visual features; and (3) the
highly informative perceptual features can be drawn out of a possibly
infinite set.

These advantages of RLVC are essentially due to the fact that the
candidate visual features are not selected only because they are
informative: They are also {\it ranked\/} according to an
information-theoretic measure inspired by decision tree
induction~\cite{Breiman:book-1984}. Such a ranking is required, as
vision-for-action tasks induce a large number of visual features (a
typical image contains about a thousand of them). This kind of
criterion that ranks features, though already considered in Variable
Resolution Grids \cite{Munos:ML2002}, seems to be new in discrete
perceptual spaces.

RLVC is defined independently of any fixed RL algorithm, which is
similar in spirit to Continuous U-Tree \cite{Uther:AAAI1998}, with the
major exception that RLVC deals with Boolean features, whereas
Continuous U-Tree works in a continuous input space. Furthermore, the
version of RLVC presented in this paper uses a variance-reduction
criterion for ranking the visual features. This criterion, though
already considered in Variable Resolution Grids, seems to be new in
discrete perceptual spaces.

%% file: compacting.tex
\section{Compacting Visual Policies}
\label{sec-compact}

As written in Section~\ref{intro-compact}, this original version of
RLVC is subject to overfitting~\cite{Jodogne:EWRL2005}. A simple
heuristic to avoid the creation of too many visual classes is simply
to bound the number of visual classes that can be refined at each
stage of the algorithm, since splitting {\it one\/} visual class
potentially has an impact on the Bellman residuals of {\it all\/} the
visual classes. In practice, we first try to split the classes that
have the most samples before considering the others, since there is
more evidence of variance reduction for the first. In our tests, we
systematically apply this heuristics. However, it is often
insufficient if taken alone.

\subsection{Equivalence Relations in Markov Decision Processes}

Since we apply an embedded RL algorithm at each stage $k$ of RLVC,
properties like the optimal value function $V^*_k(\cdot)$, the optimal
state-action value function $Q^*_k(\cdot,\cdot)$ and the optimal
control policy $\pi^*_k(\cdot)$ are known for each mapped MDP ${\cal
M}_k$. Using those properties, it is easy to define a whole range of
equivalence relations between the visual classes. For instance, given
a threshold $\varepsilon\in \Rset^+$, we list hereunder three possible
equivalence relations for a pair of visual classes $(V,V')$:
\begin{description}
\item[Optimal Value Equivalence:]~\par
$|V^*_k(V)-V^*_k(V')| \leq \varepsilon.$
\item[Optimal Policy Equivalence:]~\par
$|V^*_k(V)-Q^*_k(V', \pi^*_k(V))| \leq \varepsilon~\wedge$\\
$|V^*_k(V')-Q^*_k(V, \pi^*_k(V'))| \leq \varepsilon.$
\item[Optimal State-Action Value Equivalence:]~\par
$(\forall a\in A)\ |Q^*_k(V,a)-Q^*_k(V',a)| \leq \varepsilon.$
\end{description}

We therefore propose to modify RLVC so that, periodically, visual
classes that are equivalent with respect to one of those criteria are
merged together.  We have experimentally observed that the conjunction
of the first two criteria tends to lead to the best performance. This
way, RLVC alternatively splits and merges visual classes. The
compaction phase should not be done too often, in order to allow
exploration. To the best of our knowledge, this possibility has not
been investigated yet in the framework of adaptive-resolution methods
in reinforcement learning.

In the original version of RLVC, the visual classes correspond to the
leaves of a decision tree. When using decision trees, the aggregation
of visual classes can only be achieved by starting from the bottom of
the tree and recursively collapsing leaves, until dissimilar leaves
are found. This operation is very close to {\it post-pruning\/} in the
framework of decision trees for machine
learning~\cite{Breiman:book-1984}. In practice, this means that
classes that have similar properties, but that can only be reached
from one another by making a number of hops upwards then downwards,
are extremely unlikely to be matched. This greatly reduces the
interest of exploiting the equivalence relations.

This drawback is due to the rather limited expressiveness of decision
trees. In a decision tree, each visual class corresponds to a
conjunction of visual feature literals, which defines a path from
the root of the decision tree to one leaf. To take full advantage of
the equivalence relations, it is necessary to associate, to each
visual class, an arbitrary {\it union\/} of conjunctions of
visual features. Indeed, when exploiting the equivalence
relations, the visual classes are the result of a sequence of
conjunctions (splitting) and disjunctions (aggregation). Thus, a more
expressive data structure that would be able to represent general,
arbitrary Boolean combinations of visual features is required.
Such a data structure is introduced in the next section.

\subsection{Using Binary Decision Diagrams}

The problem of representing general Boolean functions has been
extensively studied in the field of computer-aided verification, since
they can abstract the behavior of logical electronic devices.  In
fact, a whole range of methods for representing the state space of
richer and richer domains have been developed over the last few years,
such as {\it Binary Decision Diagram\/} (BDD)~\cite{Bryant:ACM1992},
{\it Number\/} and {\it Queue Decision
Diagrams\/}~\cite{Boigelot:phd-1999}, {\it Upward Closed
Sets\/}~\cite{Delzanno:LNCS2000} and {\it Real Vector
Automata\/}~\cite{Boigelot:TOCL2003}.

In our framework, BDD is a particularly well-suited tool. It is a
acyclic graph-based symbolic representation for encoding arbitrary
Boolean functions, and has had much success in the field of
computer-aided verification~\cite{Bryant:ACM1992}. A BDD is unique
when the ordering of its variables is fixed, but different variable
orderings can lead to different sizes of the BDD, since some variables
can be discarded by the reordering process. Although the problem of
finding the optimal variable ordering is
coNP-complete~\cite{Bryant:TC1986}, automatic heuristics can in
practice find orderings that are close to optimal. This is interesting
in our case, since reducing the size of the BDD potentially discards
irrelevant variables, which correspond to removing useless visual
features.

\subsection{Modifications to RLVC}

To summarize, this extension to RLVC does not use decision trees
anymore, but assigns one BDD to each visual class. Two modifications
are to be applied to Algorithm~\ref{algo-rlvc}:
\begin{enumerate}
\item The operation of refining, with a visual feature $f$, a visual
class $V$ that is labeled by the BDD ${\cal B}(V)$, consists in
replacing $V$ by two new visual classes $V_1$ and $V_2$ such that
${\cal B}(V_1) = {\cal B}(V)\wedge f$ and ${\cal B}(V_2)={\cal
B}(V)\wedge \neg f$.

\item Given an equivalence relation, the $\mbox{\tt
post-process}({\cal C}_k)$ operation consists in merging the
equivalent visual classes. To merge a pair of visual classes
$(V_1,V_2)$, $V_1$ and $V_2$ are deleted, and a new visual class $V$
such that ${\cal B}(V) = {\cal B}(V_1) \vee {\cal B}(V_2)$ is
added. Every time a merging operation takes place, it is advised to
carry on variable reordering, to minimize the memory requirements. 
\end{enumerate}

\subsection{Experiments}
\label{sec:realistic}

\begin{figure}[t]
\begin{center}
\begin{minipage}{8cm}
  \includegraphics[width=\linewidth]{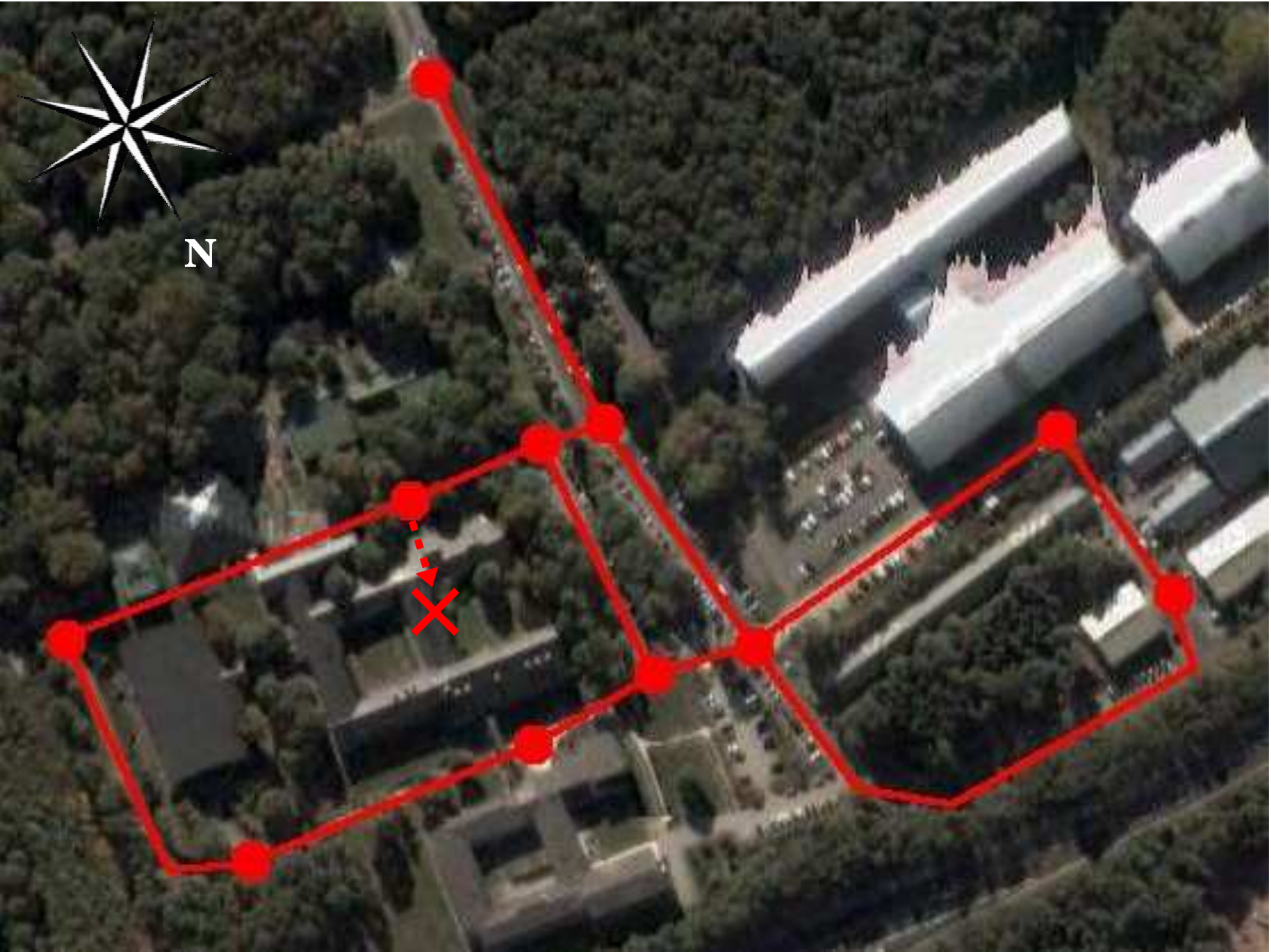}
  \flushright{\tiny (c) Google Map}
\end{minipage}
\end{center}
\caption{The Montefiore campus at Li\`ege. Red spots corresponds to
the places between which the agent moves. The agent can only follow
the links between the different spots. Its goal is to enter the
Montefiore Institute, that is labeled by a red cross, where it gets a
reward of $+100$. \label{fig-campus}}
\end{figure}

\begin{figure}[t]
\begin{center}
\begin{minipage}{14cm}
  \includegraphics[width=\linewidth]{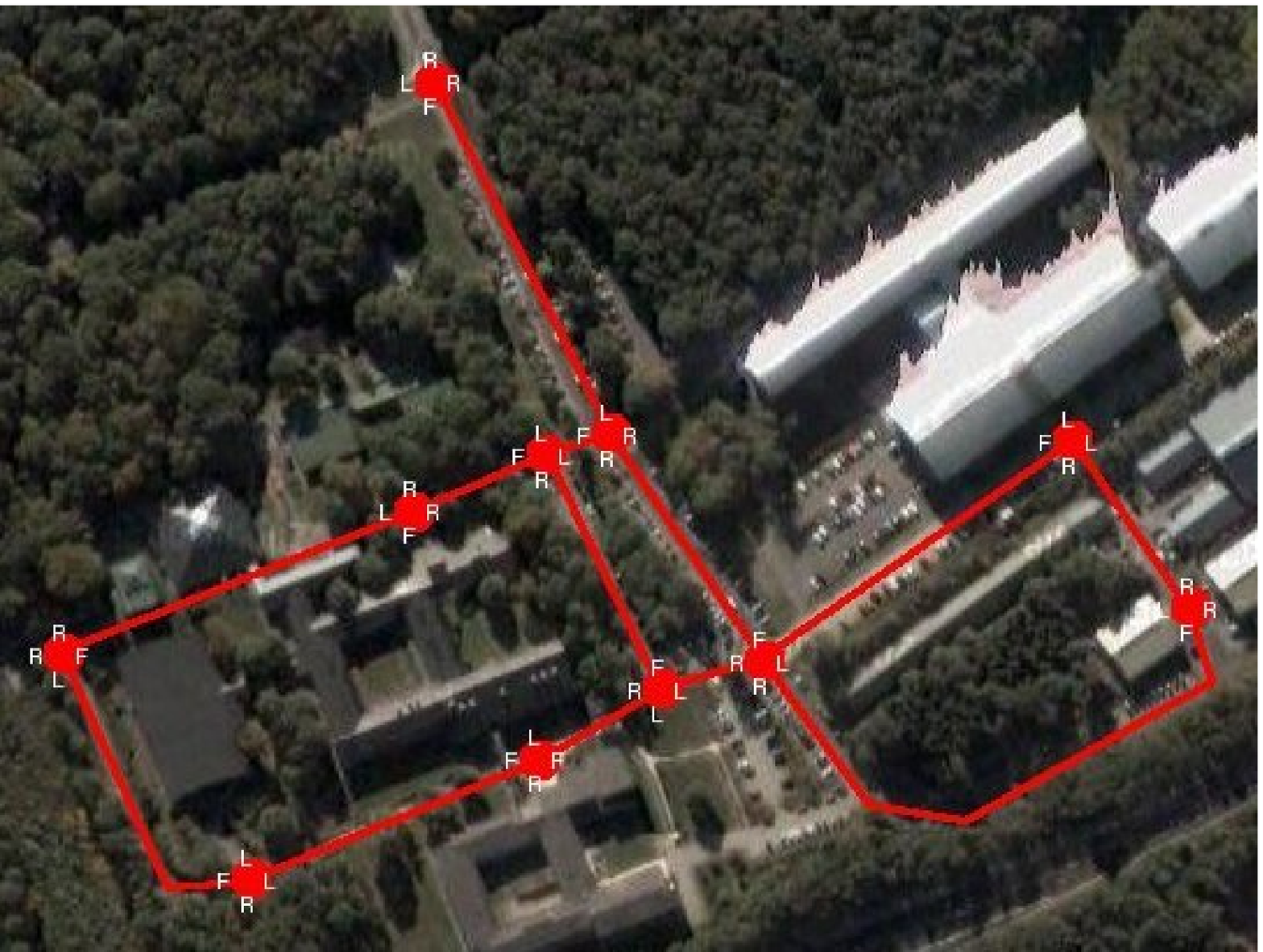}
  \flushright{\tiny (c) Google Map}
\end{minipage}
\end{center}
\caption{One of the optimal, deterministic control policies for the
Montefiore navigation task. For each state, we have indicated the
optimal action (the letter ``F'' stands for ``move forward'', ``R''
for ``turn right'' and ``L'' for ``turn left''). This policy has been
obtained by applying a standard RL algorithm to the scenario in which
the agent has direct access to the $(p,d)$
information. \label{fig-montef-policy}}
\end{figure}

\begin{figure}
\begin{center}
\begin{minipage}{6cm}
\includegraphics[width=\linewidth]{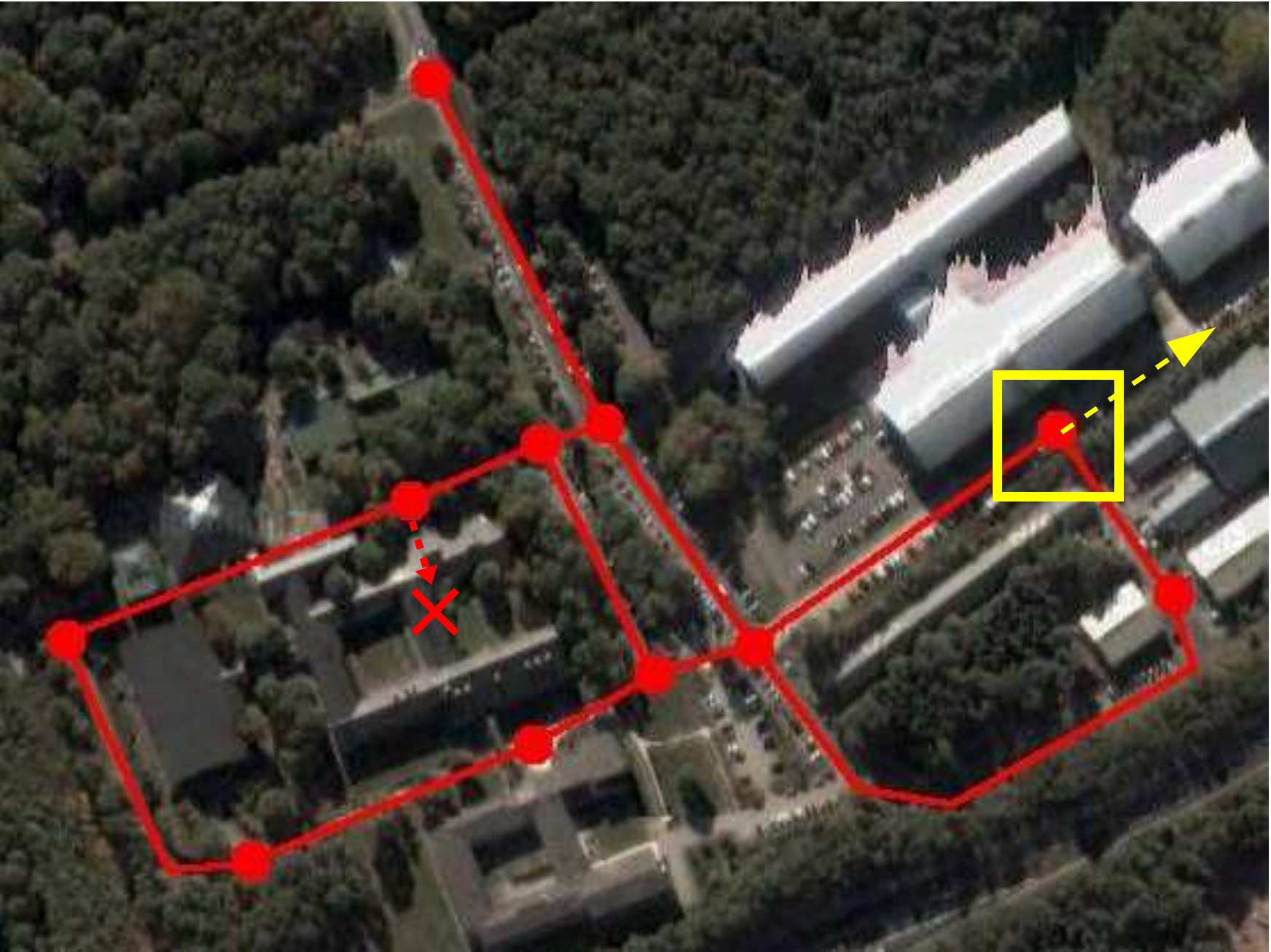} \\[-2em]
  \flushright{\tiny (c) Google Map}
\end{minipage}\\[4em]
\includegraphics[width=14cm]{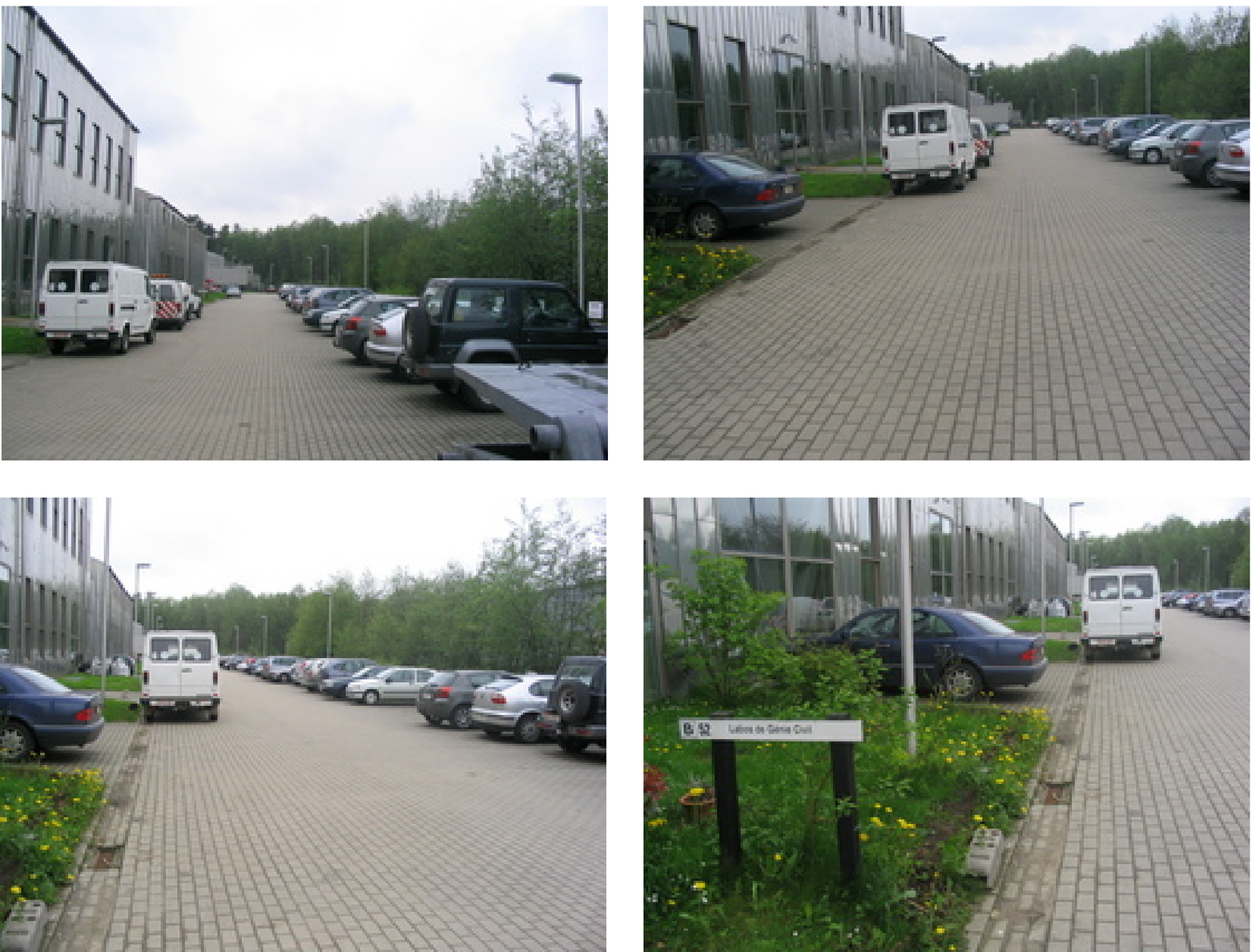}
\end{center}
\caption{The percepts of the agent. Four possible different percepts
are shown, that correspond to the location and viewing direction
marked in yellow on the top image.\label{fig-viewpoint}}
\end{figure}

We have applied the modified version of RLVC to another simulated
navigation task. In this task, the agent moves between 11 spots of the
campus of the University of Li{\`e}ge
(cf. Figure~\ref{fig-campus}). Every time the agent is at one of the
11 locations, its body can aim at four possible orientations: North,
South, West, East. The state space is therefore of size $11\times
4=44$. The agent has three possible actions: Turn left, turn right, go
forward. Its goal is to enter a specific building, where it will
obtain a reward of $+100$. Turning left or right induces a penalty of
$-5$, and moving forward, a penalty of $-10$. The discount factor
$\gamma$ is set to $0.8$. The optimal control policy is not unique:
One of them is depicted on Figure~\ref{fig-montef-policy}.

The agent does not have direct access to its position and its
orientation. Rather, it only perceives a picture of the area that is
in front of it (cf. Figure~\ref{fig-viewpoint}). Thus, the agent has
to connect an input image to the appropriate reaction without
explicitly knowing its geographical localization. For each possible
location and each possible viewing direction, a database of 24 images
of size $1024\times 768$ with significant viewpoint changes has been
collected. Those 44 databases have been randomly divided into a
learning set of 18 images and a test set of 6 images. In our
experimental setup, both versions of RLVC learn an image-to-action
mapping using interactions that only contain images from the learning
set. Images from the test set are used to assess the accuracy of the
learned visual control policies.

The SIFT keypoints have been used as visual
features~\cite{Lowe:IJCV2004}. Thresholding on a Mahalanobis distance
gave rise to a set of 13,367 distinct features. Both versions of RLVC
have been applied on a static database of 10,000 interactions that has
been collected using a fully randomized exploration policy. The same
database is used throughout the entire algorithm, and this database
only contains images that belong to the learning set.

\begin{figure}[t]
\begin{center}
\includegraphics[width=7.5cm]{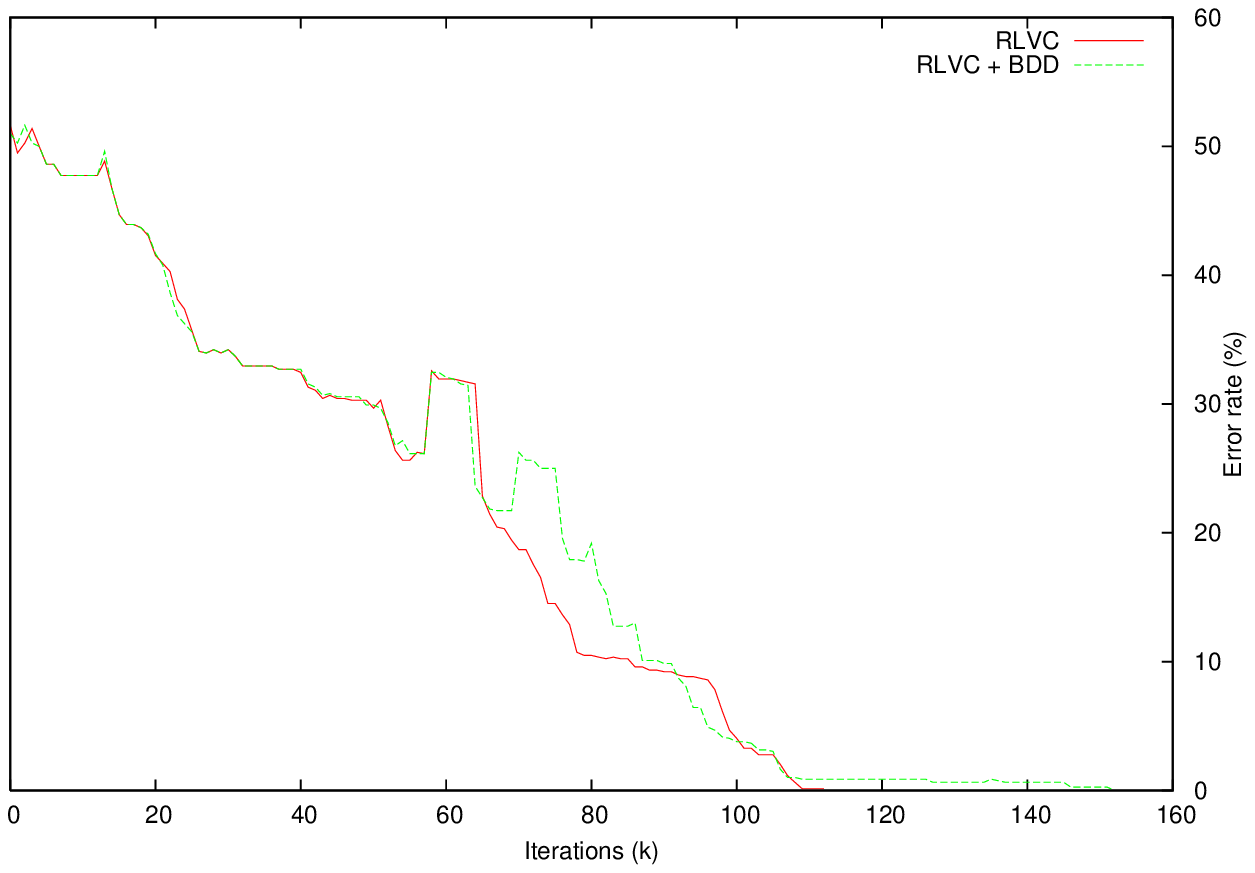}~~
\includegraphics[width=7.5cm]{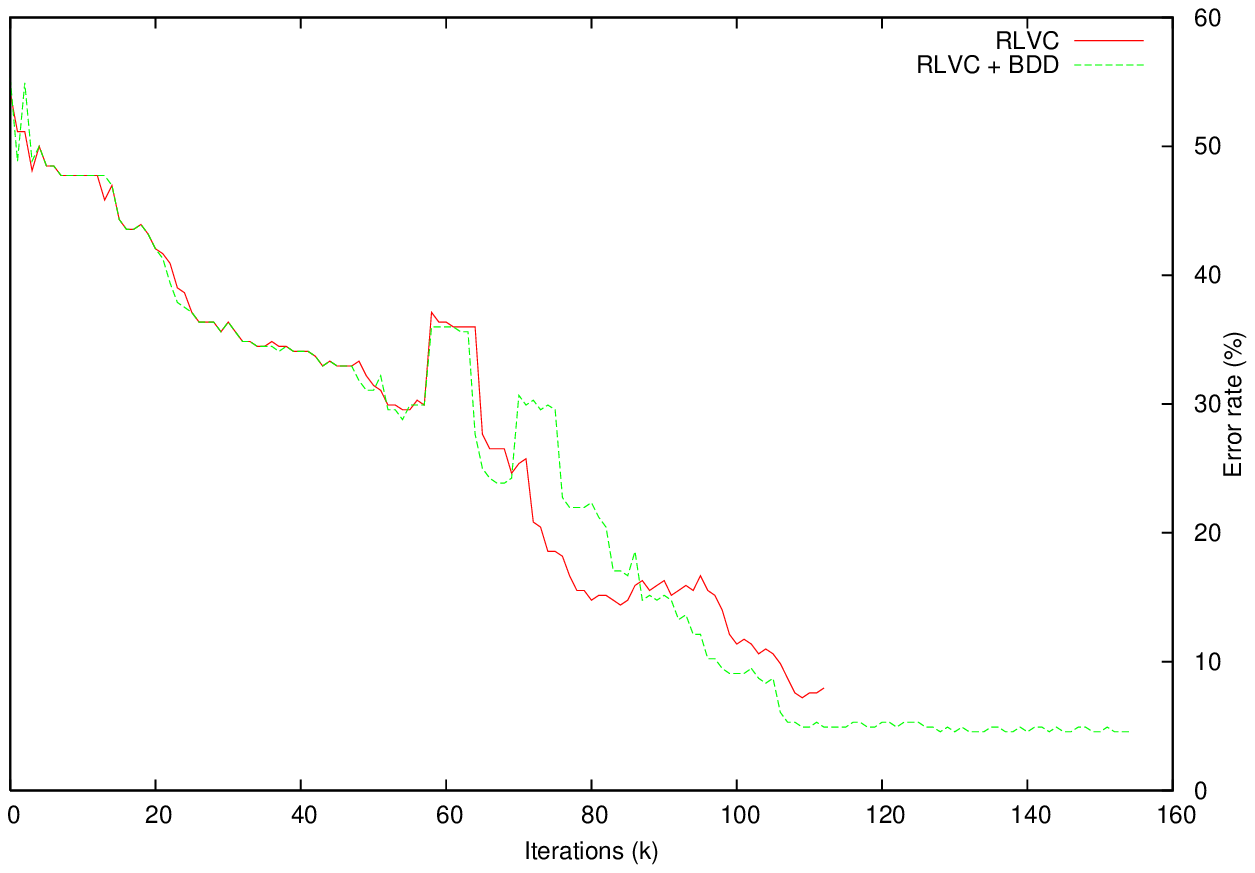}
\end{center}
\caption{Comparison of the error rates between the basic and extended
versions of RLVC. The error of the computed policy as a function of
the step counter $k$ on the images of the learning set (resp. test
set) is reported on the left of the figure (resp. on the
right).\label{fig-error-rate}}
\end{figure}

The results of the basic version of RLVC and of the version that is
extended by BDDs are reported in Figures~\ref{fig-error-rate} and
\ref{fig-numbers}. The original version of RLVC has identified 281
visual classes by selecting 264 SIFT features. The error rate on the
computed visual policy (i.e.~the proportion of sub-optimal decisions
when the agent is presented all the possible stimuli) was $0.1\%$ on
the learning set and $8\%$ when the images of the test set are used,
with respect to the optimal policy when the agent has direct access to
its position and viewing direction.

The modified version RLVC was then applied, with one compacting stage
every 10 steps. The results are clearly superior. There is no error on
the learning set anymore, while the error rate on the test set is
$4.5\%$. The number of selected features is reduced to
171. Furthermore, the resulting number of visual classes becomes 59,
instead of 281. Thus, there is a large improvement in the
generalization abilities, as well as a reduction of the number of
visual classes and selected features. Interestingly enough, the number
of visual classes (59) is very close to the number of physical states
(44), which tends to indicate that the algorithm starts to learn a
physical interpretation for its percepts.

To summarize, compacting visual policies is probably a required step
to deal with realistic visual tasks, if an iterative splitting process
is applied. The price to pay is of course a higher computational cost.
Future work will focus on a theoretical justification of the used
equivalence relations. This implies bridging the gap with the theory
of MDP minimization~\cite{Givan:AI2003}.

%% file: spatial.tex
\section{Learning Spatial Relationships}
\label{sec-spatial}

As motivated in the Introduction (Section \ref{intro-spatial}), we
propose to extend RLVC by constructing a hierarchy of spatial
arrangements of individual point
features~\cite{Jodogne:CVPRWL2005}. The idea of learning models of
spatial combinations of features takes its roots in the seminal paper
by Fischler and Elschlager~\citeyear{Fischler:TC1973} about pictorial
structures, which are collections of rigid parts arranged in
deformable configurations. This idea has become increasingly popular
in the computer vision community over the 90's, and has led to a large
literature about the modeling and the detection of
objects~\cite{Amit:TPAMI1996,Burl:CVPR1996,Forsyth:SCGCV1999}.
Crandall and Huttenlocher~\citeyear{Crandall:ECCV2006} provide
pointers to recent resources. Among such recent techniques, Scalzo and
Piater~\citeyear{Scalzo:ICPR2006} propose to build a probabilistic
hierarchy of visual features that is represented through an acyclic
graph. They detect the presence of such a model through Nonparametric
Belief Propagation~\cite{Sudderth:CVPR2003}. Other graphical models
have been proposed for representing articulated structures, such as
pictorial
structures~\cite{Felzenszwalb:IJCV05,Kumar:BMVC2004}. Similarly, the
constellation model represents objects by parts, each modeled in terms
of both shape and appearance by Gaussian probability density
functions~\cite{Perona:CVPR2003}.

\begin{figure}[t]
\begin{center}
\includegraphics[width=7.5cm]{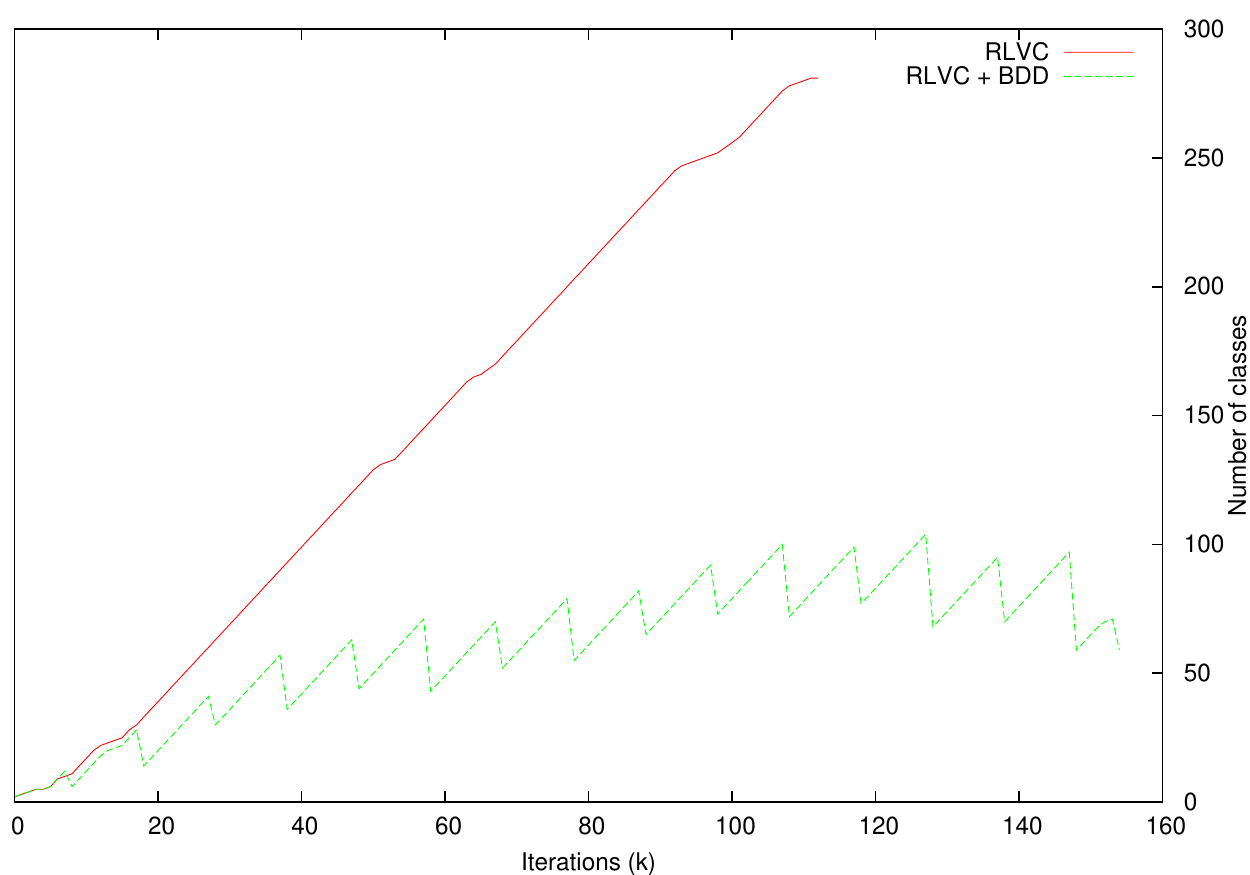}~~
\includegraphics[width=7.5cm]{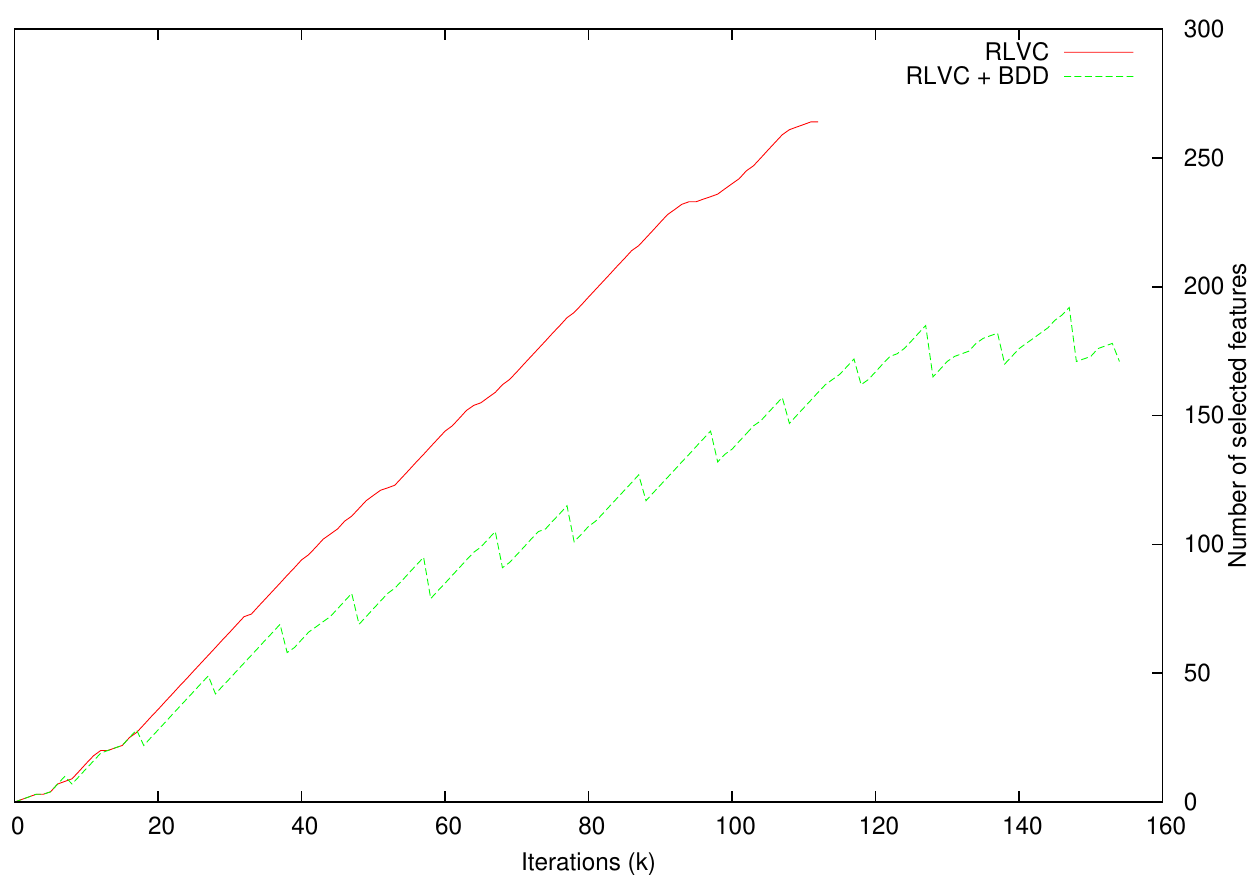}
\end{center}
\caption{Comparison of the number of generated classes and selected
visual features between the basic and extended versions of RLVC. The
number of visual classes (resp. selected features) as a function of
the step counter $k$ is plotted on the left of the figure (resp. on
the right).\label{fig-numbers}}
\end{figure}

Our work contrasts with these approaches in that the generation of
so-called {\it composite features\/} is driven by the task to be
solved. This should be distinguished from the techniques for
unsupervised learning of composite features, since the additional
information that is embedded inside the reinforcement signal drives
the generation of composite features by focusing the exploration on
task-relevant spatial arrangements.

In this extension of RLVC, a hierarchy of visual features is built
simultaneously with the image classifier. As soon as no sufficiently
informative visual feature can be extracted, the algorithm tries to
combine two visual features in order to construct a higher level of
abstraction, which is hopefully more distinctive and more robust to
noise. This extension to RLVC assumes the co-existence of two
different kinds of visual features:
\begin{description}
\item[Primitive Features:] They correspond to the individual point
features, i.e.~to the local-appearance descriptors
(cf. Section~\ref{local-appearance}). 
\item[Composite Features:] They consist of spatial combinations of
lower-level visual features.  There is no {\it a priori\/} bound on
the maximal height of the hierarchy.  Therefore, a composite feature
can be potentially combined with a primitive feature, or with a
composite feature.
\end{description}

\subsection{Detection of Visual Features}

A natural way to represent such a hierarchy is to use a directed
acyclic graph $G=(V,E)$, in which each vertex $v\in V$ corresponds to
a visual feature, and in which each edge $(v,v')\in E$ models the fact
that $v'$ is a part of the composite feature $v$. Thus, $G$ must be
{\it binary\/}, i.e.~any vertex should have either no child, or
exactly two children. The set $V_P$ of the leaves of $G$ corresponds
to the set of primitive features, while the set $V_C$ of its internal
vertexes represents the set of composite features.

Each leaf vertex $v_P\in V_P$ is annotated with a local descriptor
$D(v_P)$. Similarly, each internal vertex $v_C\in V_C$ is annotated
with constraints on the relative position between its parts. In this
work, we consider only constraints on the distances between the
constituent visual features of the composite features, and we assume
that they should be distributed according to a Gaussian law ${\cal
G}(\mu,\sigma)$ of mean $\mu$ and standard deviation
$\sigma$. Evidently, richer constraints could be used, such as taking
the relative orientation or the scaling factor between the constituent
features into consideration, which would require the use of
multivariate Gaussians.

More precisely, let $v_C$ be a composite feature, the parts of which
are $v_1$ and $v_2$. In order to trigger the detection of $v_C$ in an
image $s$, there should be an occurrence of $v_1$ and an occurrence of
$v_2$ in $s$ such that their relative Euclidean distance has a
sufficient likelihood $\nu$ of being generated by a Gaussian of mean
$\mu(v_C)$ and standard deviation $\sigma(v_C)$. To ensure symmetry,
the location of the composite feature is then taken as the midpoint
between the locations of $v_1$ and $v_2$.

The occurrences of a visual feature $v$ in a percept $s$ can be found
using the recursive Algorithm~\ref{algo-detect}.  Of course, at steps
6 and 7 of Algorithm~\ref{algo-selection}, the test ``does $s_t$
exhibit $v$?'' can be rewritten as a function of
Algorithm~\ref{algo-detect}, by checking if $\mbox{\tt occurrences}(v,
s_t)\not=\emptyset$.

\begin{algorithm}[p]
\caption{--- Detecting Composite Features\label{algo-detect}}
\begin{algorithmic}[1]
 \NAME{$\mbox{\tt occurrences}(v, s)$}
 \IF{$v$ is primitive}
  \STATE {\bf return} \{$(x,y) \mid$ $(x,y)$ is an interest
  point of $s$, the local descriptor of which corresponds to $D(v)$\}
 \ELSE
  \STATE $O\leftarrow \{\}$
  \STATE $O_1 \leftarrow \mbox{\tt occurrences}(\mbox{\tt subfeature}_1(v), s)$
  \STATE $O_2 \leftarrow \mbox{\tt occurrences}(\mbox{\tt subfeature}_2(v), s)$
  \FORALL{$(x_1,y_1)\in O_1$ and $(x_2,y_2)\in O_2$}
   \STATE $d\leftarrow \sqrt{(x_2-x_1)^2 + (y_2-y_1)^2}$
   \IF{${\cal G}(d-\mu(v),\sigma(v)) \geq \nu$}
    \STATE $O\leftarrow O\cup \{((x_1+x_2)/2, (y_1+y_2)/2)\}$
   \ENDIF
  \ENDFOR
  \STATE {\bf return} $O$
 \ENDIF
 \ENDNAME
\end{algorithmic}
\end{algorithm}

\subsection{Generation of Composite Features}

The cornerstone of this extension to RLVC is the way of generating
composite features. The general idea behind our algorithm is to
accumulate statistical evidence from the relative positions of the
detected visual features in order to find ``conspicuous coincidences''
of visual features. This is done by providing a more evolved
implementation of $\mbox{\tt generator}(s_1,\ldots,s_n)$ than the one
of Algorithm~\ref{algo-simple-generator}.

\subsubsection{Identifying Spatial Relations}

We first extract the set $F$ of all the (primitive or composite)
features that occur within the set of provided images
$\{s_1,\ldots,s_n\}$:
\begin{equation}
F = \left\{v\in V \mid (\exists i)~s_i \mbox{~exhibits~} v\right\}.
\end{equation}
We identify the pairs of visual features the occurrences of which are
highly correlated within the set of provided images
$\{s_1,\ldots,s_n\}$. This simply amounts to counting the number of
co-occurrences for each pair of features in $F$, then only keeping the
pairs the corresponding count of which exceeds a fixed threshold.

Let now $v_1$ and $v_2$ be two features that are highly correlated. A
search for a meaningful spatial relationship between $v_1$ and $v_2$
is then carried out in the images $\{s_1,\ldots,s_n\}$ that contain
occurrences of both $v_1$ and $v_2$. For each such co-occurrence, we
accumulate in a set $\Lambda$ the distances between the corresponding
occurrences of $v_1$ and $v_2$.  Finally, a clustering algorithm is
applied on the distribution $\Lambda$ in order to detect typical
distances between $v_1$ and $v_2$. For the purpose of our experiments,
we have used {\it hierarchical clustering\/}~\cite{Jain:ACM99}. For
each cluster, a Gaussian is fitted by estimating a mean value $\mu$
and a standard deviation $\sigma$. Finally, a new composite feature
$v_C$ is introduced in the feature hierarchy, that has $v_1$ and $v_2$
as parts and such that $\mu(v_C)=\mu$ and $\sigma(v_C)=\sigma$.

In summary, in Algorithm~\ref{algo-rlvc}, we replace the call to
Algorithm~\ref{algo-simple-generator} by a call to
Algorithm~\ref{algo-composite-generator}. 


\begin{algorithm}[p]
\caption{--- Generation of Composite
 Features\label{algo-composite-generator}}
\begin{algorithmic}[1]
\NAME{$\mbox{\tt generator}(\{s_1,\ldots,s_n\})$}
\STATE $F = \left\{v\in V \mid (\exists i)~s_i \mbox{~exhibits~} v\right\}$
\STATE $F' = \{\}$
\FORALL{$(v_1,v_2)\in F\times F$}
\IF{enough co-occurrences of $v_1$ and $v_2$ in $\{s_1,\ldots,s_n\}$}
\STATE $\Lambda \leftarrow \{\}$
\FORALL{$i\in \{1,\ldots,n\}$}
\FORALL{occurrences $(x_1,y_1)$ of $v_1$ in $s_i$}
\FORALL{occurrences $(x_2,y_2)$ of $v_2$ in $s_i$}
\STATE $\Lambda \leftarrow \Lambda \cup \{\sqrt{(x_2-x_1)^2 + (y_2-y_1)^2}\}$
\ENDFOR
\ENDFOR
\ENDFOR
\STATE Apply a clustering algorithm on $\Lambda$
\FOR{each cluster $C=\{d_1,\ldots d_m\}$ in $\Lambda$}
\STATE $\mu=\mean(C)$
\STATE $\sigma=\stddev(C)$
\STATE Add to $F'$ a composite feature $v_C$ composed of $v_1$ and $v_2$,
annotated with a mean $\mu$ and a standard deviation $\sigma$
\ENDFOR
\ENDIF
\ENDFOR
\STATE {\bf return} $F'$
\ENDNAME
\end{algorithmic}
\end{algorithm}

\subsubsection{Feature Validation}

Algorithm~\ref{algo-composite-generator} can generate several
composite features for a given visual class $V_{k,i}$. However, at the
end of Algorithm~\ref{algo-selection}, at most one generated composite
feature is to be kept. It is important to notice that the performance
of the clustering method is not critical for our purpose. Indeed,
irrelevant spatial combinations are automatically discarded, thanks to
the variance-reduction criterion of the feature selection
component. In fact, the reinforcement signal helps to direct the
search for a good feature, which is an advantage over unsupervised
methods of building feature hierarchies.

\subsection{Experiments}

We demonstrate the efficacy of our algorithms on a version of the
classical ``Car on the Hill'' control problem \cite{Moore:ML1995},
where the position and velocity information is presented to the agent
visually.

In this episodic task, a car (modeled by a mass point) is riding
without friction on a hill, the shape of which is defined by the
function:
$$H(p) = \left\{ \begin{array}{ll}
p^2+p & \mbox{if~} p < 0, \\
p / \sqrt{1+5p^2} & \mbox{if~} p \geq 0.
\end{array}\right.$$
The goal of the agent is to reach as fast as possible the top of the
hill, i.e.~a location such that $p\geq 1$. At the top of the hill, the
agent obtains a reward of 100. The car can thrust left or right with
an acceleration of $\pm 4$ Newtons. However, because of gravity, this
acceleration is insufficient for the agent to reach the top of the
hill by always thrusting toward the right. Rather, the agent has to go
left for while, hence acquiring potential energy by going up the left
side of the hill, before thrusting rightward. There are two more
constraints: The agent is not allowed to reach locations such that
$p<-1$, and a velocity greater than 3 in absolute value leads to the
destruction of the car.

\begin{figure}[t]
\begin{center}
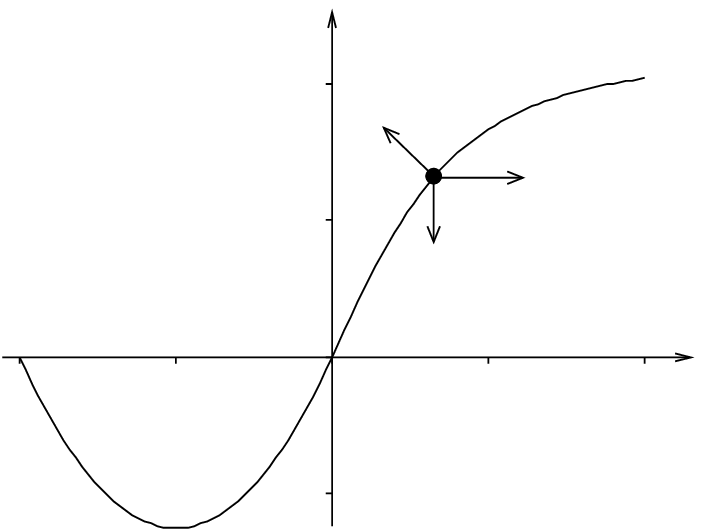
\end{center}
\caption{The ``Car on the Hill'' control problem.\label{car-on-the-hill}}
\end{figure}

\subsubsection{Formal Definition of the Task}

Formally, the set of possible actions is $A=\{-4,4\}$, while the state
space is
$S=\{(p,s) \mid |p|\leq 1 \wedge |s|\leq 3\}.$
The system has the following continuous-time dynamics:
\begin{eqnarray*}
\dot{p} & = & s \\[-.5em] \dot{s} & = &
\frac{a}{M\sqrt{1+H'(p)^2}} - \frac{g H'(p)}{1+H'(p)^2},
\end{eqnarray*}
where $a\in A$ is the thrust acceleration, $H'(p)$ is the first
derivative of $H(p)$, $M=1$ is the mass of the car, and $g=9.81$ is
the acceleration due to gravity. These continuous-time dynamics are
approximated by the following discrete-time state update rule:
\begin{eqnarray*}
s_{t+1} & = & s_t + h\dot{p}_t + h^2 \dot{s}_t / 2 \\
s_{t+1} & = & \dot{p}_t + h \dot{s}_t,
\end{eqnarray*}
where $h=0.1$ is the integration time step. The reinforcement signal
is defined through this expression:
$${\cal R}((s_t,s_t),a) = \left\{\begin{array}{ll}
100 & \mbox{if~} s_{t+1}\geq 1 \wedge |s_{t+1}| \leq 3, \\
0 & \mbox{otherwise}.
\end{array}\right.$$
In our setup, the discount factor $\gamma$ was set to
$0.75$.

This definition is actually a mix of two coexistent formulations of
the ``Car on the Hill'' task~\cite{Ernst:ECML2003,Moore:ML1995}. The
major differences with the initial formulation of the
problem~\cite{Moore:ML1995} is that the set of possible actions is
discrete, and that the goal is at the top of the hill (rather than on
a given area of the hill), just like in the definition from Ernst et
al.~\citeyear{Ernst:ECML2003}. To ensure the existence of an
interesting solution, the velocity is required to remain less than 3
(instead of 2), and the integration time step is set to $h=0.1$
(instead of $0.01$).

\subsubsection{Inputs of the Agent}

In previous work~\cite{Moore:ML1995,Ernst:ECML2003}, the agent was
always assumed to have direct access to a numerical measure of its
position and velocity. The only exception is Gordon's work in which a
visual, low-resolution representation of the global scene is given to
the agent~\cite{Gordon:ICML1995}. In our experimental setup, the agent
is provided with two cameras, one looking at the ground underneath,
the second at a velocity gauge. This way, the agent cannot directly
know its current position and velocity, but has to suitably interpret
its visual inputs to derive them.

\begin{figure}[t]
\begin{center}
\includegraphics[width=9cm]{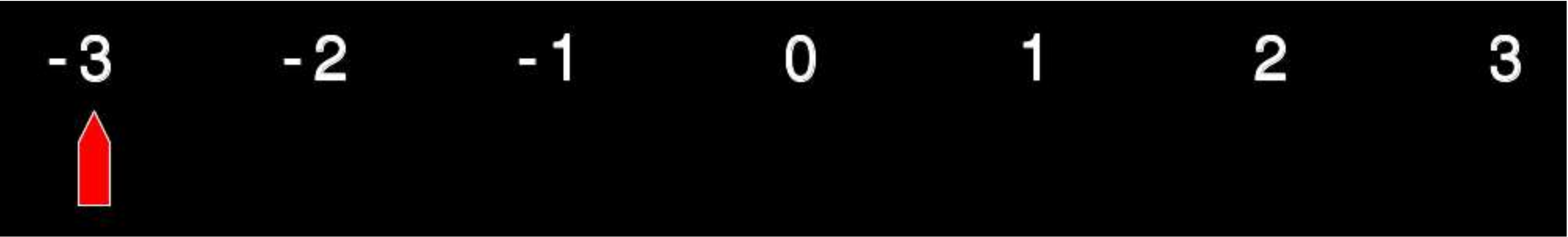} \\[1em]
\includegraphics[width=9cm]{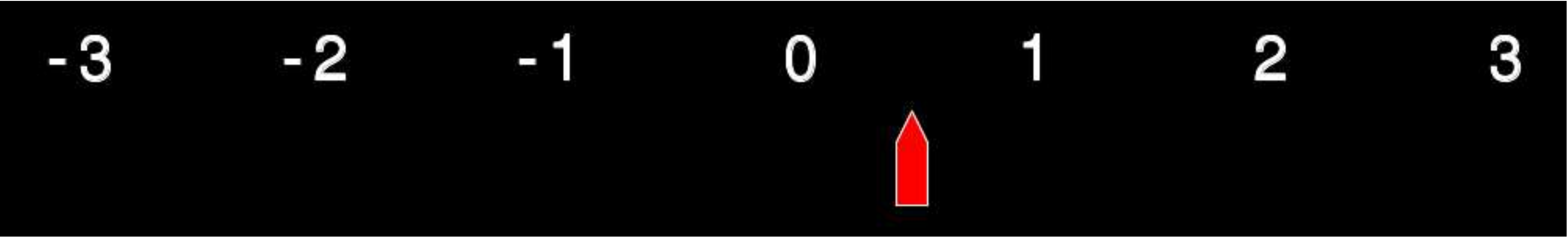} \\[1em]
\includegraphics[width=9cm]{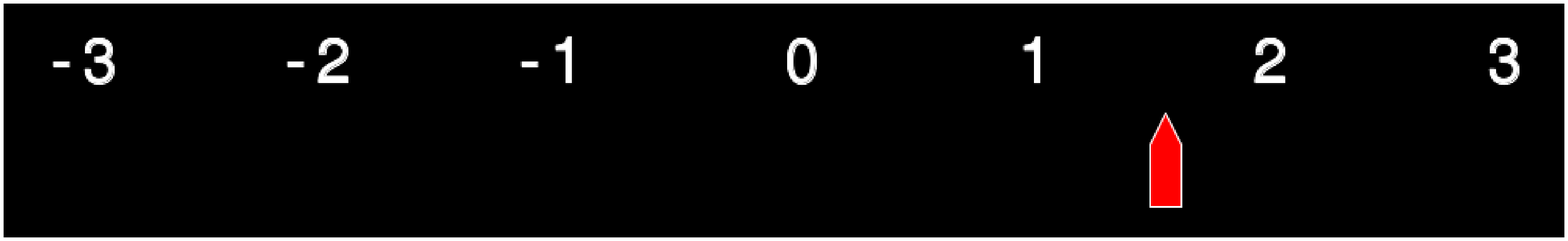} \\[1em]
(a)\\[1em]
\includegraphics[width=15cm]{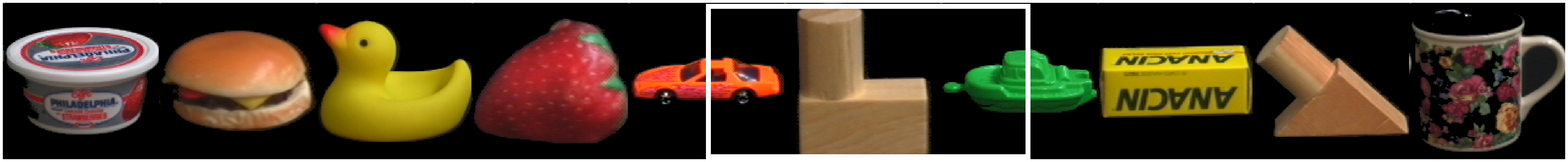} \\
(b)
\end{center}
\caption{(a) Visual percepts corresponding to pictures of the velocity
gauge when $s=-3$, $s=0.5$ and $s=1.5$. (b) Visual percepts returned
by the position sensor. The region framed with a white rectangle
corresponds to the portion of the ground that is returned by the
sensor when $p=0.1$. This portion slides back and forth as the agent
moves.
\label{fig-sensors}}
\end{figure}

Some examples of the pictures the sensors can return are presented in
Figure~\ref{fig-sensors}. The ground is carpeted with a color image of
$1280\times 128$ pixels that is a montage of pictures from the
COIL-100 database~\cite{COIL100-96}. It is very important to notice
that using individual point features is insufficient for solving this
task, since the set of features in the pictures of the velocity gauge
are always the same. To know its velocity, the agent has to generate
composite features sensitive to the distance of the primitive features
on the cursor with respect to the primitive features on the digits.

\subsubsection{Results}

In this experimental setup, we used color differential invariants
\cite{Gouet:CBAIVL2001} as primitive features. Among all the
possible visual inputs (both for the position and the velocity
sensors), there were 88 different primitive features. The entire image
of the ground includes 142 interest points, whereas the images of the
velocity gauge include about 20 interest points.

\begin{figure}[t]
\centerline{\includegraphics[width=10cm]{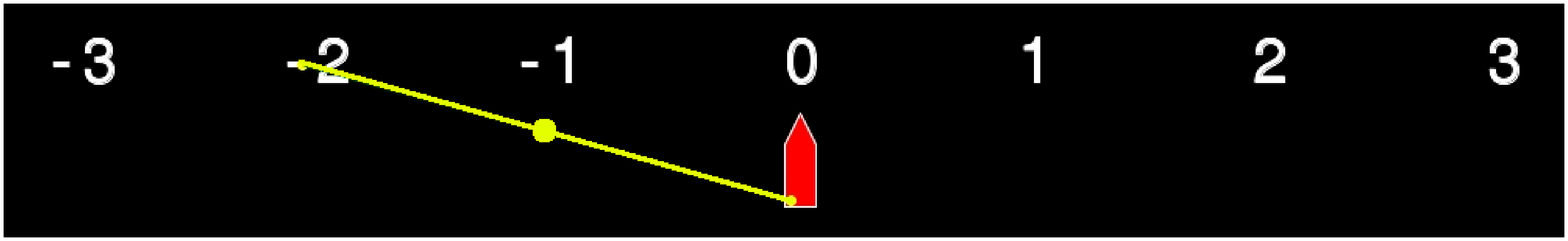}}~\\[.5em]
\centerline{\includegraphics[width=10cm]{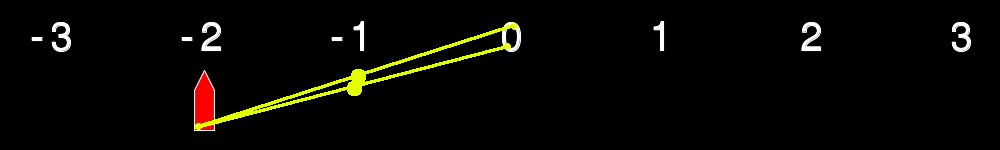}}
\caption{Two composite features that were generated, in yellow. The
primitive features of which they are composed are marked in yellow.
The first feature triggers for velocities around 0, whereas the second
triggers around $-2$.\label{fig-composite}}
\end{figure}

The output of RLVC is a decision tree that defines 157 visual
classes. Each internal node of this tree tests the presence of one
visual feature, taken from a set of 91 distinct, highly discriminant
features selected by RLVC. Among the 91 selected visual features,
there were 56 primitive and 26 composite features.  Two examples of
composites features that were selected by RLVC are depicted in
Figure~\ref{fig-composite}.  The computation stopped after $k=38$
refinement steps in Algorithm~\ref{algo-rlvc}.

\begin{figure}
\begin{center}
\includegraphics[width=8cm]{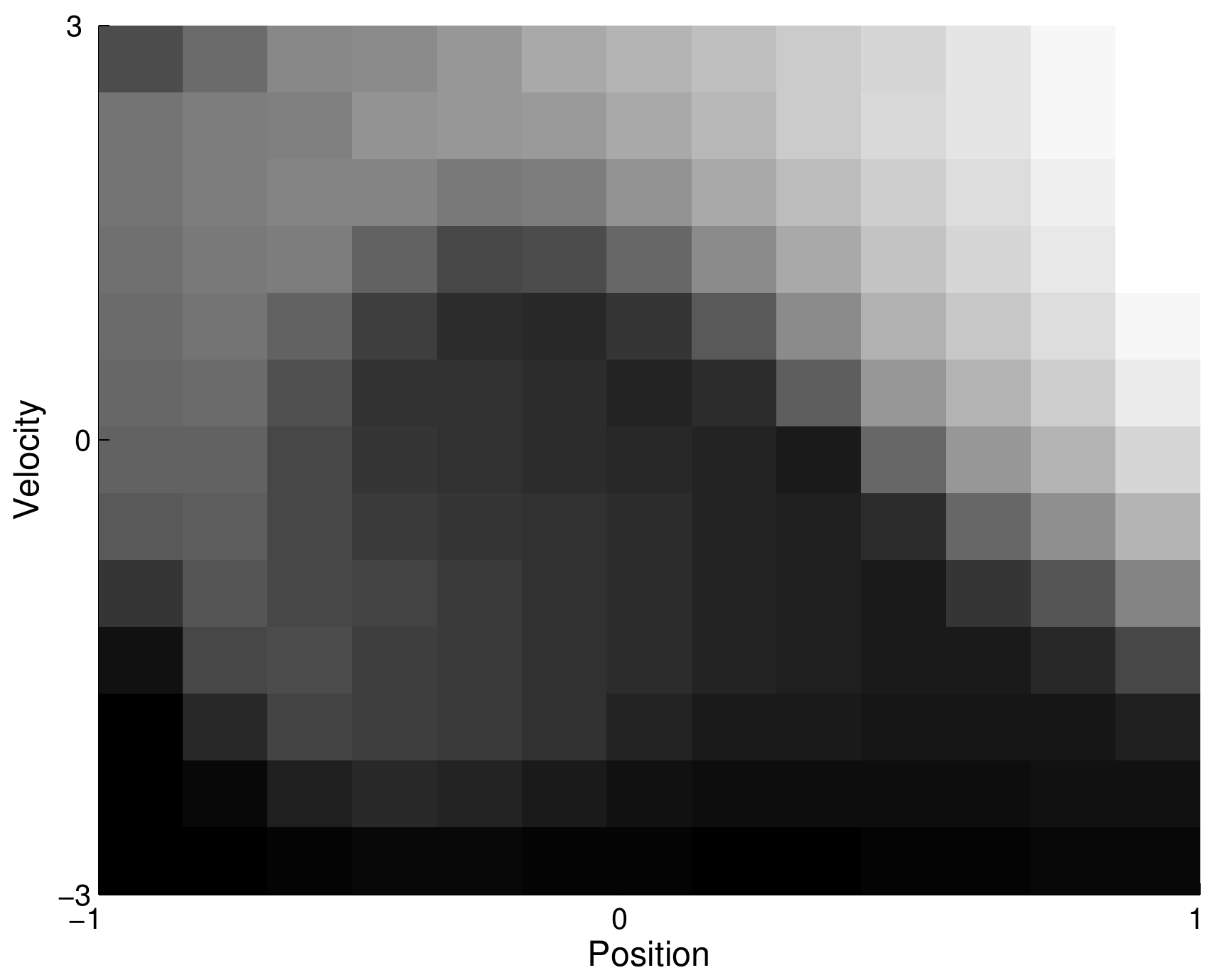} \\ (a)\\[1em]
\includegraphics[width=8cm]{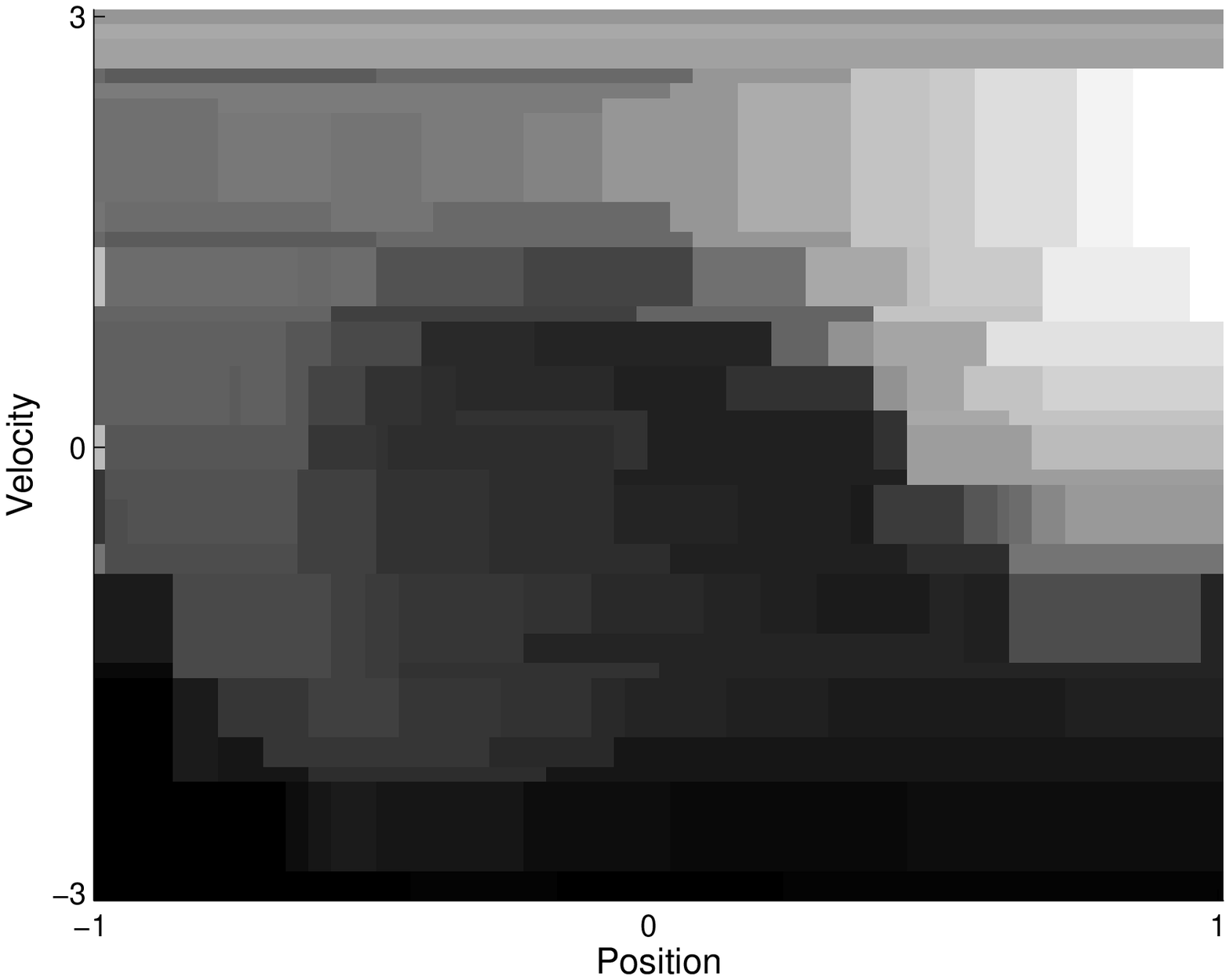} \\ (b)
\end{center}
\caption{(a) The optimal value function, when the agent has a direct
access to its current $(p,s)$ state, and when the input space is
discretized in a $13\times 13$ grid. The brighter the location, the
greater its value. (b) The value function obtained by
RLVC.\label{fig-car-values}}
\end{figure}

\begin{figure}[p]
\begin{center}
\includegraphics[width=10cm]{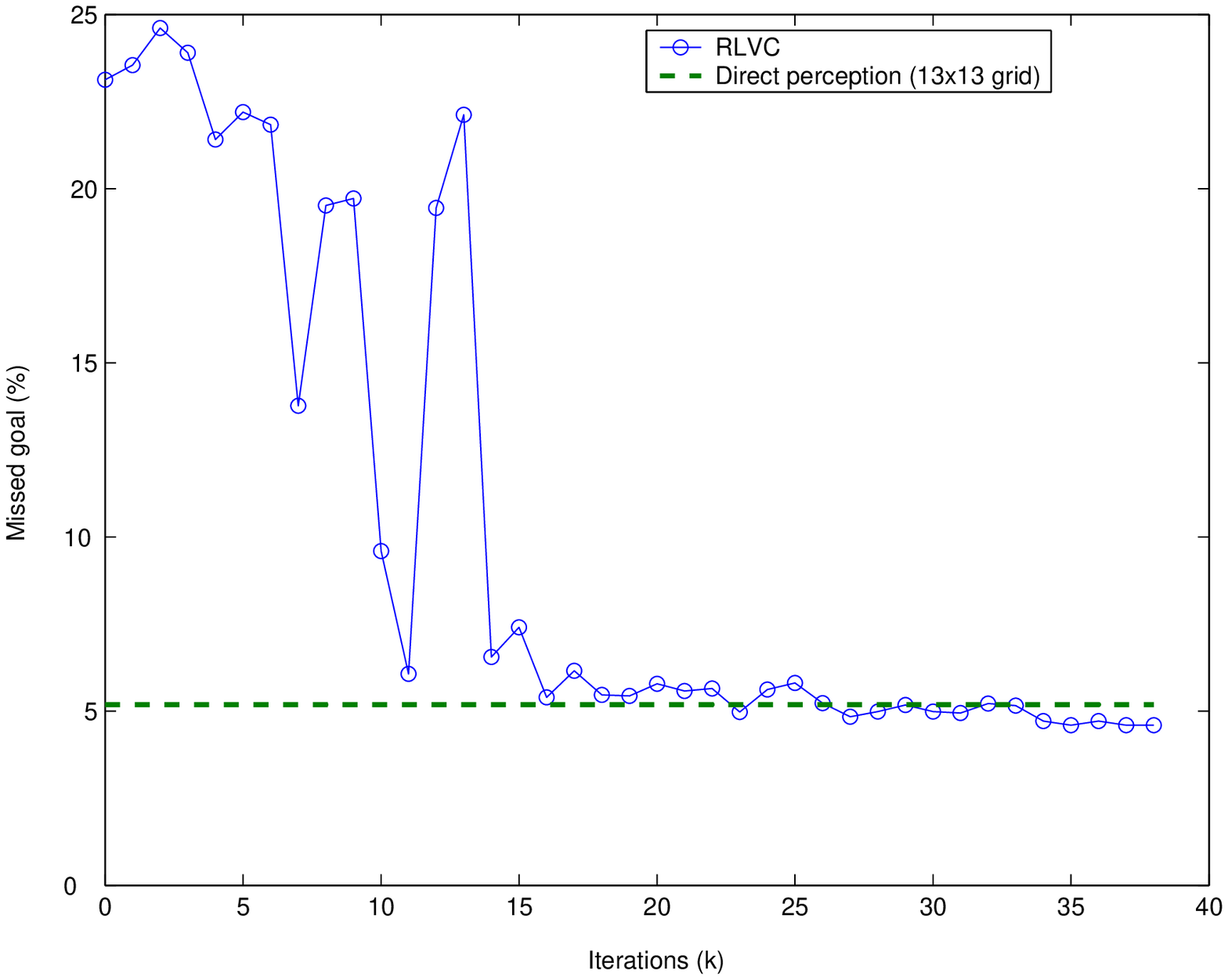}
\end{center}
\caption{Evolution of the number of times the goal was missed over the
iterations of RLVC.\label{fig-missed}}
\end{figure}

\begin{figure}[p]
\begin{center}
\includegraphics[width=10cm]{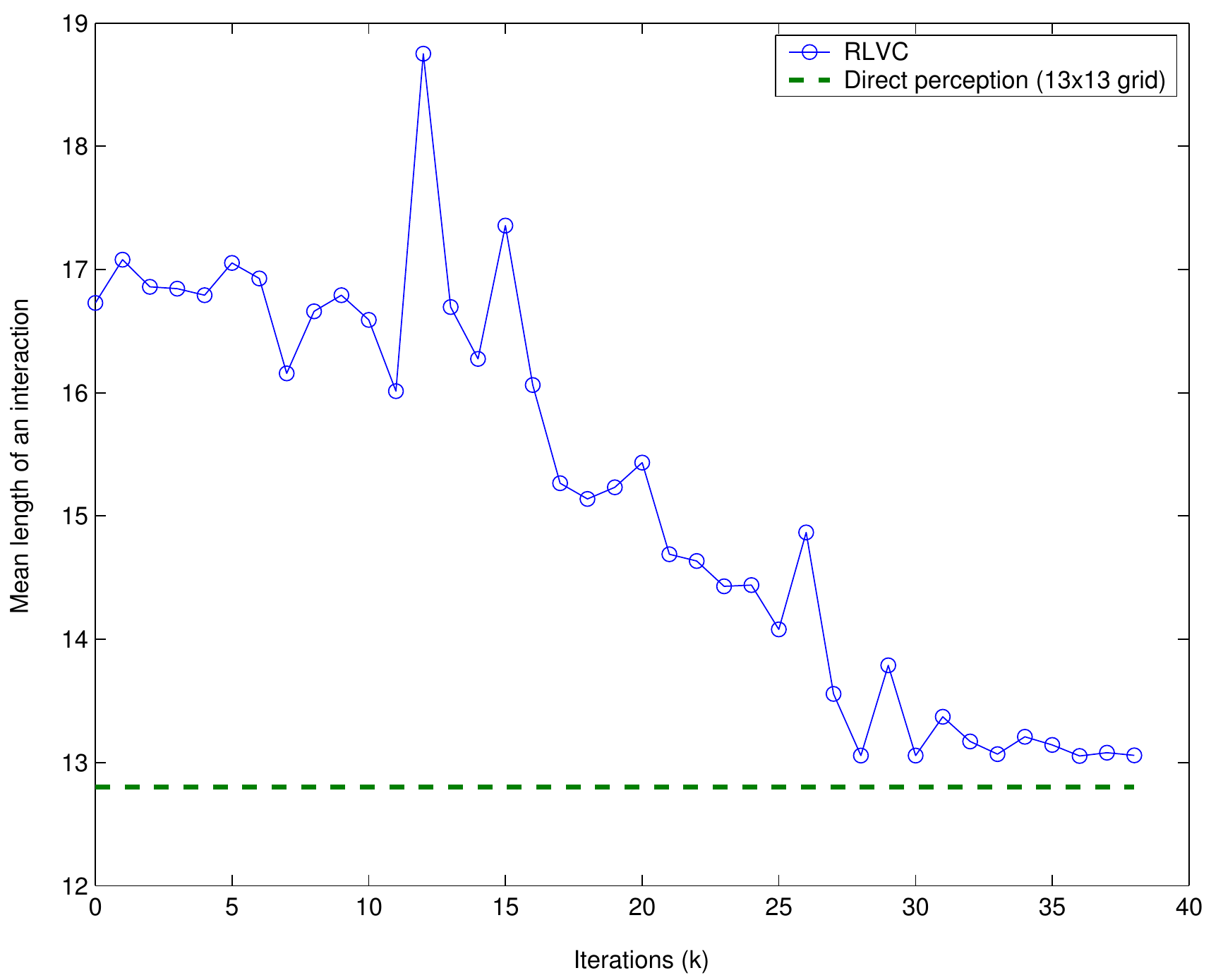}
\end{center}
\caption{Evolution of the mean lengths of the successful trials over
the iterations of RLVC.\label{fig-mean}}
\end{figure}

To show the efficacy of our method, we compare its performance with
the scenario in which the agent has a direct perception of its current
$(p,s)$ state. In the latter scenario, the state space was discretized
in a grid of $13\times 13$ cells. The number 13 was chosen since it
approximately corresponds to the square root of 157, the number of
visual classes that were produced by RLVC. This way, RL is provided an
equivalent number of perceptual classes in the two
scenarios. Figure~\ref{fig-car-values} compares the optimal value
function of the direct-perception problem with the one obtained
through RLVC. Here also, the two pictures are very similar, which
indicates the soundness of our approach.

We have also evaluated the performance of the optimal image-to-action
mapping
\begin{equation}
  \pi^*=\argmax_{a\in A}Q^*((p,s),a)
\end{equation}
obtained through RLVC.  For this purpose, the agent was placed
randomly on the hill, with an initial velocity of 0. Then, it used the
mapping $\pi^*$ to choose an action, until it reached a final state. A
set of 10,000 such trials were carried out at each step $k$ of
Algorithm~\ref{algo-rlvc}.  Figure~\ref{fig-missed} compares the
proportion of trials that missed the goal (either because of leaving
the hill on the left, or because of acquiring a too high velocity) in
RLVC and in the direct-perception problem. When $k$ became greater
than 27, the proportion of missed trials was always smaller in RLVC
than in the direct-perception problem. This advantage in favor of RLVC
is due to the adaptive nature of its
discretization. Figure~\ref{fig-mean} compares the mean lengths of the
successful trials. The mean length of RLVC trials clearly converges to
that of the direct-perception trials, while staying slightly larger.

To conclude, RLVC achieves a performance close to the
direct-perception scenario. However, the mapping built by RLVC
directly links visual percepts to the appropriate actions, without
considering explicitly the physical variables.


%% file: figs/car-on-the-hill.pstex_t
\begin{picture}(0,0)%
\includegraphics{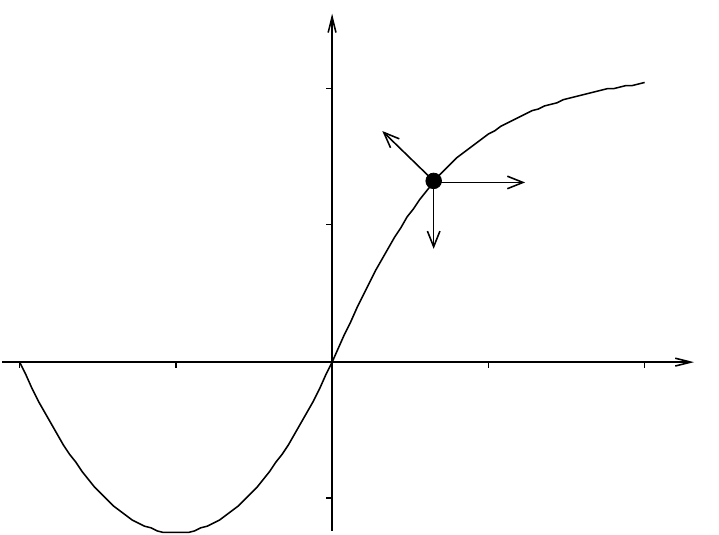}%
\end{picture}%
\setlength{\unitlength}{1973sp}%
\begingroup\makeatletter\ifx\SetFigFont\undefined%
\gdef\SetFigFont#1#2#3#4#5{%
  \reset@font\fontsize{#1}{#2pt}%
  \fontfamily{#3}\fontseries{#4}\fontshape{#5}%
  \selectfont}%
\fi\endgroup%
\begin{picture}(6877,5113)(939,-5348)
\put(2576,-3586){\makebox(0,0)[lb]{\smash{{\SetFigFont{9}{10.8}{\familydefault}{\mddefault}{\updefault}{\color[rgb]{0,0,0}$-.5$}%
}}}}
\put(1076,-3586){\makebox(0,0)[lb]{\smash{{\SetFigFont{9}{10.8}{\familydefault}{\mddefault}{\updefault}{\color[rgb]{0,0,0}$-1$}%
}}}}
\put(5576,-3586){\makebox(0,0)[lb]{\smash{{\SetFigFont{9}{10.8}{\familydefault}{\mddefault}{\updefault}{\color[rgb]{0,0,0}$.5$}%
}}}}
\put(7076,-3586){\makebox(0,0)[lb]{\smash{{\SetFigFont{9}{10.8}{\familydefault}{\mddefault}{\updefault}{\color[rgb]{0,0,0}$1$}%
}}}}
\put(4801,-2986){\makebox(0,0)[lb]{\smash{{\SetFigFont{12}{14.4}{\familydefault}{\mddefault}{\updefault}{\color[rgb]{0,0,0}$mg$}%
}}}}
\put(6151,-2056){\makebox(0,0)[lb]{\smash{{\SetFigFont{12}{14.4}{\familydefault}{\mddefault}{\updefault}{\color[rgb]{0,0,0}$u$}%
}}}}
\put(7801,-3811){\makebox(0,0)[lb]{\smash{{\SetFigFont{12}{14.4}{\familydefault}{\mddefault}{\updefault}{\color[rgb]{0,0,0}$p$}%
}}}}
\put(4351,-1336){\makebox(0,0)[lb]{\smash{{\SetFigFont{11}{13.2}{\familydefault}{\mddefault}{\updefault}{\color[rgb]{0,0,0}$N$}%
}}}}
\put(3991,-2461){\makebox(0,0)[rb]{\smash{{\SetFigFont{9}{10.8}{\familydefault}{\mddefault}{\updefault}{\color[rgb]{0,0,0}$0.2$}%
}}}}
\put(3991,-1156){\makebox(0,0)[rb]{\smash{{\SetFigFont{9}{10.8}{\familydefault}{\mddefault}{\updefault}{\color[rgb]{0,0,0}$0.4$}%
}}}}
\put(4201,-5086){\makebox(0,0)[lb]{\smash{{\SetFigFont{9}{10.8}{\familydefault}{\mddefault}{\updefault}{\color[rgb]{0,0,0}$-0.2$}%
}}}}
\put(3901,-586){\makebox(0,0)[rb]{\smash{{\SetFigFont{12}{14.4}{\familydefault}{\mddefault}{\updefault}{\color[rgb]{0,0,0}$H(p)$}%
}}}}
\end{picture}%

%% file: conclusions.tex
\section{Summary}

This paper introduces {\it Reinforcement Learning of Visual Classes\/}
(RLVC). RLVC is designed to learn mappings that directly connect
visual stimuli to output actions that are optimal for the surrounding
environment. The framework of RLVC is general, in the sense that it
can be applied to any problem that can be formulated as a Markov
decision problem.

The learning process behind our algorithms is closed-loop and
flexible. The agent takes lessons from its interactions with the
environment, according to the purposive vision paradigm. RLVC focuses
the attention of an embedded reinforcement learning algorithm on
highly informative and robust parts of the inputs by testing the
presence or absence of local descriptors at the interest points of the
input images. The relevant visual features are incrementally selected
in a sequence of attempts to remove perceptual aliasing: The
discretization process targets zero Bellman residuals and is inspired
from supervised learning algorithms for building decision trees.  Our
algorithms are defined independently of any interest point
detector~\cite{Schmid:IJCV2000} and of any local description
technique~\cite{Mikolajczyk:CVPR2003}. The user may choose these two
components as he sees fit.

Techniques for fighting overfitting in RLVC have also been proposed.
The idea is to aggregate visual classes that share similar properties
with respect to the theory of Dynamic Programming. Interestingly, this
process enhances the generalization abilities of the learned
image-to-action mapping, and reduces the number of visual classes that
are built.

Finally, an extension of RLVC is introduced that allows the
closed-loop, interactive and purposive learning of a hierarchy of
geometrical combinations of visual features. This is to contrast to
most of the prior work on the topic, that uses either a supervised or
unsupervised
framework~\cite{Piater:phd-2001,Fergus:CVPR03,Bouchard:CVPR05,Scalzo:ICPR2006}. Besides
the novelty of the approach, we have shown its practical value in
visual control tasks in which the information provided by the
individual point features alone is insufficient for solving the
task. Indeed, spatial combinations of visual features are more
informative and more robust to noise.

\section{Future Work}

The area of applications of RLVC is wide, since nowadays robotic
agents are often equipped with CCD sensors.  Future research includes
the demonstration of the applicability of our algorithms in a reactive
robotic application, such as grasping objects by combining visual and
haptic feedback~\cite{Coelho:RAS2001}. This necessitates the extension
of our techniques to continuous action spaces, for which no fully
satisfactory solutions exist to date.  RLVC could also be potentially
be applied to Human-Computer Interaction, as the actions need not be
physical actions.

The closed-loop learning of a hierarchy of visual feature also raises
interesting research directions. For example, the combination of RLVC
with techniques for disambiguating between aliased percepts using a
short-term memory~\cite{McCallum:phd-1996} could solve visual tasks in
which the percepts of the agent alone do not provide enough
information for solving the task. Likewise, the unsupervised learning
of other kinds of geometrical models~\cite{Felzenszwalb:IJCV05} could
potentially be embedded in RLVC. On the other hand, spatial
relationships do not currently take into consideration the relative
angles between the parts of a composite feature. Doing so would
further increase the discriminative power of the composite features,
but requires non-trivial techniques for clustering in circular
domains.

\section*{Acknowledgments}

The authors thank the associate editor Thorsten Joachims and the three
anonymous reviewers for their many suggestions for improving the
quality of the manuscript. S\'ebastien Jodogne gratefully acknowledge
the financial support of the Belgian National Fund for Scientific
Research (FNRS).